\documentclass[letterpaper, 10 pt, conference]{ieeeconf}

\IEEEoverridecommandlockouts                              

\usepackage{graphics} 
\usepackage{epsfig} 
\usepackage{times} 
\usepackage{amsmath} 
\usepackage{amssymb}  
\usepackage{amsmath, bm}
\usepackage{float}
\usepackage{subfigure}
\usepackage{threeparttable}
\usepackage{tabularx}
\usepackage{multirow}
\usepackage{booktabs}
\usepackage[linesnumbered,ruled,vlined]{algorithm2e}
\usepackage{algorithmicx}
\usepackage{url}
\usepackage[normalem]{ulem}

\usepackage{color}
\newcolumntype{Y}{>{\centering\arraybackslash}m{0.7cm}}

\newcommand{\magenta}[1]{\textcolor{magenta}{#1}}

\begin{document}

\title{\LARGE \bf
Learning to Dexterously Pick or Separate Tangled-Prone Objects\\for Industrial Bin Picking
}

\author{Xinyi Zhang$^{1*}$, 
Yukiyasu Domae$^{2}$, Weiwei Wan$^{1}$ and Kensuke Harada$^{1,2}$
\thanks{*Correspond to: {\tt\small xinyiz0931@gmail.com}}
\thanks{$^{1}$Graduate School of Engineering Science, Osaka University, Japan}%
\thanks{$^{2}$Industrial Cyber Physical Systems Research Center, National Institute of Advanced Industrial Science and Technology (AIST), Japan}
}

\maketitle

\thispagestyle{plain}
\pagestyle{plain}

\begin{abstract}
    
    Industrial bin picking for tangled-prone objects requires the robot to either pick up untangled objects or perform separation manipulation when the bin contains no isolated objects. The robot must be able to flexibly perform appropriate actions based on the current observation. It is challenging due to high occlusion in the clutter, elusive entanglement phenomena, and the need for skilled manipulation planning. In this paper, we propose an autonomous, effective and general approach for picking up tangled-prone objects for industrial bin picking. First, we learn PickNet - a network that maps the visual observation to pixel-wise possibilities of picking isolated objects or separating tangled objects and infers the corresponding grasp. Then, we propose two effective separation strategies: Dropping the entangled objects into a buffer bin to reduce the degree of entanglement; Pulling to separate the entangled objects in the buffer bin planned by PullNet - a network that predicts position and direction for pulling from visual input. To efficiently collect data for training PickNet and PullNet, we embrace the self-supervised learning paradigm using an algorithmic supervisor in a physics simulator. Real-world experiments show that our policy can dexterously pick up tangled-prone objects with success rates of 90\%. We further demonstrate the generalization of our policy by picking a set of unseen objects. Supplementary material, code, and videos can be found at \magenta{\url{https://xinyiz0931.github.io/tangle}}. 

\end{abstract}


\section{Introduction}\label{sec:intro}

    Bin picking is a valuable task in manufacturing to automate the assembly process. It deploys robots to pick necessary objects from disorganized bins, rather than relying on human workers to arrange the objects or using a large number of part feeders. Existing studies have tackled some challenges in bin picking such as planning grasps under rich contact between the robot's hand and the objects in dense clutter \cite{domae2014fast,buchholz2014combining,matsumura2018learning,tachikake2020learning} and visual processing heavy occluded scenes \cite{liu2012fast,choi2012voting,harada2013probabilistic,harada2018experiments,yang2021probabilistic}. However, objects with complex shapes still remain challenging for bin picking. These objects easily get entangled when randomly placed in a bin, making it difficult for the robot to pick up a single object at a time. It poses challenges in perception since the robot must be able to distinguish the isolated and potentially tangled objects in a cluttered environment. Manipulation is also challenging for planning effective and general separation motions due to the complexity of entanglement estimation and real-world executions. 

\begin{figure}[t] 
    \centering
    \includegraphics[width=\linewidth]{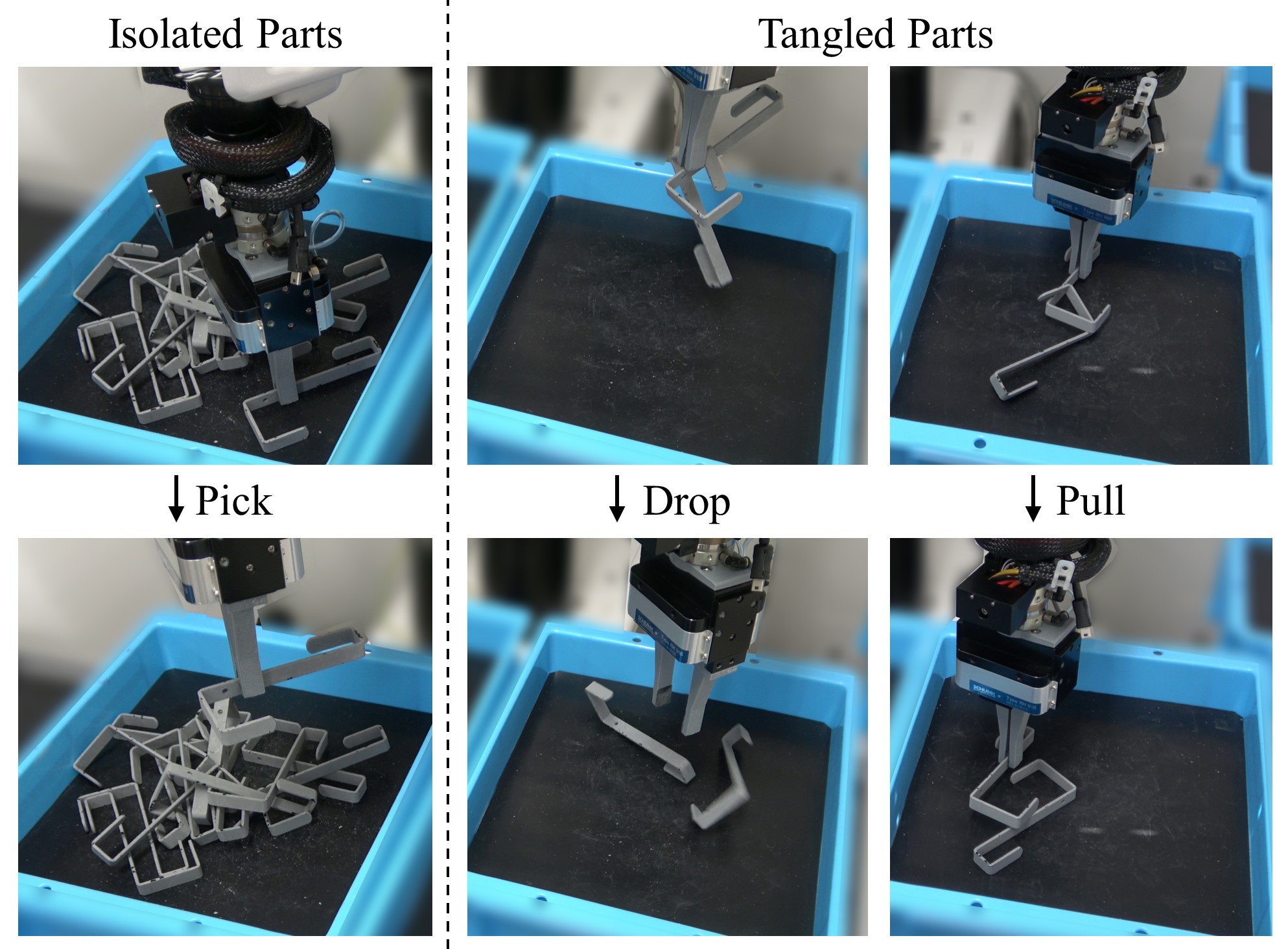}    
    \caption{Our policy learns to flexibly pick or separate tangled-prone objects for bin picking. The robot can search the untangled objects in the bin and pick them up. In cases where all objects are entangled, the robot can separate them by dropping them into another bin. Additionally, the robot can also perform pulling actions to disentangle the objects. }
    \label{fig:teaser} 
\end{figure}
    
    Prior works have addressed this problem by avoiding grasping potentially entangled objects \cite{matsumura2019learning,zhang2021topological}. However, these approaches use partial visual observation or simple geometrical features such as edges, making it challenging to be adopted in dense clutter. Other studies estimates the pose of object and evaluate the entanglement level for each object \cite{moosmann2021separating,moosmann2022transfer}. Such a paradigm relies on the full knowledge of the objects and may suffer from cumulative perception errors due to heavy occlusion or self-occlusion of an individual complex-shaped object. Other studies utilize force and torque sensors to classify if the robot grasps multiple objects \cite{moreira2016assessment}. The entanglement detection naturally leads to the necessity for exploring separation strategies. Studies have proposed tilting the gripper to discard the entangled objects \cite{leao2020detecting} or dragging the entangled object out of the clutter \cite{moosmann2021separating}. However, these object-specific strategies require prior knowledge of objects and may be insufficient for objects with different geometries. Additionally, the aforementioned learning-based approaches rely on simulated supervision \cite{moosmann2021separating} or verifying the entanglement by simulated execution \cite{matsumura2019learning}. They do not provide any general criteria for entanglement in cluttered environments. 
    
    To address these challenges, we present a novel bin-picking system that leverages self-supervised learning to flexibly and efficiently pick or separate various complex-shaped objects: 

\begin{itemize}
    \item We propose PickNet, which learns to map the visual observations of the unstructured bin to affordance maps that indicate the pixel-wise possibilities of potential actions: \textbf{pick}ing isolated objects or separating entangled objects. Our policy then selects the corresponding action with the highest action possibility. The network is trained with the idea that the untangled objects tend to present a complete contour in clutter, making it more interpretable than black-box classifiers or using insufficient object features. 
    
    \item We propose two efficient separation motion primitives to cope with different entanglement levels. The first motion is to \textbf{drop} the tangled objects into a buffer bin after grasping. Dropping can dynamically untangle the objects by utilizing the interactions with the environments instead of directly performing motions in dense clutter. It acts as an initial separation strategy to reduce the degree of entanglement and is suitable for a wide range of objects. The second motion is to \textbf{pull} the target object out of the entanglement. The robot can simultaneously pull and transport the objects when the degree of entanglement is rather lower, increasing action efficiency compared with dropping. We propose PullNet to infer the position and direction for pulling from visual observations. 
    
    \item We train PickNet and PullNet using synthetic data collected in a self-supervised manner. An algorithmic supervisor is used to estimate the entanglement state and increase the efficiency of the collection process. 
\end{itemize}

    Fig. \ref{fig:teaser} shows the proposed actions in our system. The contributions of this work are five-fold. (1) We propose a bin-picking system for tangled-prone objects that enlarges the accuracy, efficiency, flexibility and generalization. (2) We learn PickNet to distinguish untangled or tangled objects in clutter and infer the appropriate actions for them. (3) We propose two novel and efficient motion primitives for separating entangled objects: dropping and pulling. (4) We learn PullNet to infer the pulling actions without object models. (5) We develop an algorithm for simulated self-supervised data collection. We demonstrate the effectiveness of our method using both simulated and real-world experiments with an average success rate of 90\%. We also test our method on unseen objects and shows impressive results. 

\begin{figure*}[t]
\vspace{0.2cm}
    \centering
    \includegraphics[width=\linewidth]{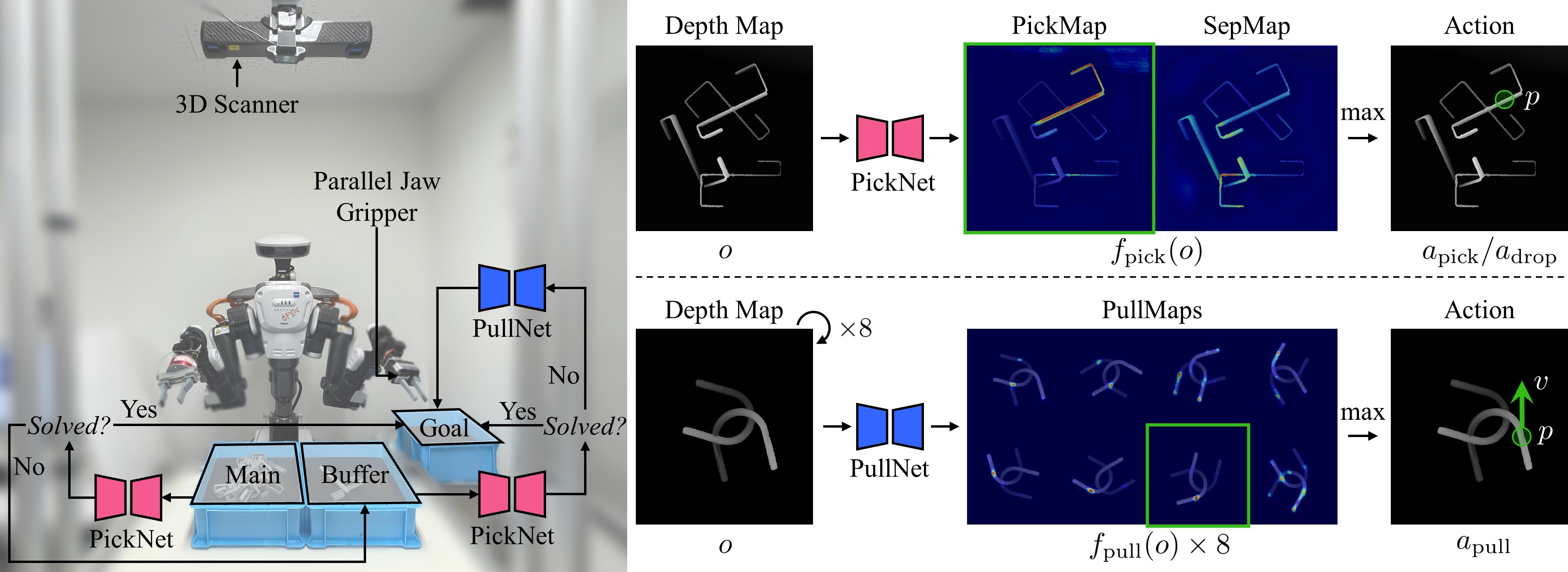}
    \caption{Overview of our policy. The robot first uses PickNet to search untangled objects in the main bin and transport them to the goal bin. If such objects do not exist, the robot grasps the entangled objects, drops them in a buffer bin and uses PickNet to check if the separation succeeds. The robot then transports the isolated objects to the goal bin or separates the tangled objects by pulling, which is inferred by PullNet. Given a depth image as input, PickNet predicts two affordance maps representing the pixel-wise possibilities of picking and separating. We rotate the depth image by eight orientations denoting eight pulling directions and feed it to PullNet. The pulling action is determined by the affordance map that yields the highest score. }
    \label{fig:overview}
\end{figure*}   

\section{Related Work}\label{sec:rw}

\subsection{Industrial Bin Picking}

    Industrial bin picking has been developed for decades. Prior works have primarily focused on model-based approaches such as 3D or 6D pose identification \cite{liu2012fast,choi2012voting,yang2021probabilistic} and grasp planning \cite{buchholz2014combining,harada2013probabilistic}. The model-free approach can directly detect grasps without object models. Domae et al. \cite{domae2014fast} proposed to plan grasps considering collisions between the gripper and the objects from a single depth image. Several studies leverage deep learning to mitigate the challenges from the complex physical phenomenon \cite{mahler2017learning,matsumura2018learning}. To further improve the robustness of bin picking, Domae et al. \cite{domae2020robotic} proposed a pipeline system for isolation, regrasping and kitting. However, there still remain challenges with objects that tend to get entangled when randomly placed in a bin. Matsumura et al. \cite{matsumura2019learning} proposed a learning-based approach to grasp avoiding potentially tangled objects. Zhang et al. \cite{zhang2021topological} generated a feature map to represent the entanglement information from a depth image. These approaches focus on searching and grasping untangled objects and are insufficient for cases where all the objects are entangled in the bin. To address this problem, Leao et al. \cite{leao2020detecting} proposed a method to pick up soft tubes by fitting shape primitives to clutter. Moosmann et al. \cite{moosmann2021separating} proposed to estimate the 6D pose of the target and then leverage reinforcement learning to plan separation motion. However, these approaches require prior knowledge of the object shape or model. On the contrary, our system does not require any prior knowledge of objects to flexibly pick isolated objects and separate entangled objects. 

\subsection{Object Singulation}

    Object singulation refers to separating an individual object in cluttered environments. Specific strategies for object singulation are required for achieving different tasks. Firstly, non-prehensile manipulation, such as pushing, is utilized to increase the grasp access when the objects are tightly placed or near the bin walls. Zeng et al. \cite{zeng2018learning} proposed to learn the synergies between pushing and grasping to create enough space for grasping in the clutter. Danielczuk et al. \cite{danielczuk2019mechanical} proposed to learn pushing policies to singulate the target object for future grasping in bin picking. Although pushing is useful for singulating daily objects or simple-shaped objects, some industrial objects face another challenge where they tend to get entangled. Studies address this problem by utilizing specifically crafted singulation strategies through model-based \cite{moosmann2021separating} or model-free approaches \cite{zhang2022learning}. Combining the advantages of these strategies, we propose two separation actions that can be planned without object models: a post-grasping action to drop the entangled objects and an action during grasping to pull out the entangled target. 
    
\subsection{Action Affordance Learning}
    
    Action affordance can be used for encoding the action with perception. For instance, grasp affordance can be learned by predicting a pixel-wise heatmap mapped with the observation where each pixel indicates the possibilities of the grasp success \cite{morrison2018closing}. Other studies also leverage action affordance for various tasks such as perceiving the 3D spatial structure of visual input for pick-and-place task \cite{zeng2021transporter}, predicting the keypoints associated with visual input for cable untangling \cite{grannen2020untangling} or inferring both position and direction for manipulating articulated objects \cite{gadre2021act}. The direction of the applied action can be encoded by rotating the input image for the inference \cite{zeng2018learning,gadre2021act}. In this paper, we employ action affordance learning in industrial bin picking to infer different actions (picking, dropping, pulling) for entangled objects. 


\section{Problem Statement}

    Let $o$ denote the depth image of the clutter, $(q,\theta)$ denote a grasp with 4 degrees of freedom, where $q \in \mathbb{R}^3,\theta \in \mathbb{R}$ is the position and orientation of the gripper about the vertical axis to the workspace.  The grasp pixel $p \in \mathbb{R}^2$ is inferred by our policy from the depth image $o$ and then transformed to a 3-D location $q$ for execution. We then leverage the method in \cite{domae2014fast} to compute the collision-free grasp orientation $\theta$ by convoluting the depth image $o$ with the gripper model. We parameterize the action $a$ with three motion behaviors: 
    
\begin{itemize}
    \item Picking: $a_\text{pick}=(q,\theta)$. The robot executes a grasp centered at $q$ oriented $\theta$, lifts in a vertically upward direction, and transports the objects to the goal bin
    \item Dropping: $a_\text{drop}=(q,\theta)$. The robot grasps at $q$ with an orientation of $\theta$ and drop the objects into the buffer bin.
    \item Pulling: $a_\text{pull}=(q,\theta,u)$. The robot executes a grasp at $(q,\theta)$, pulls along $u \in \mathbb{R}^3$, and transports it to the goal bin. Our policy produces a 2-D pulling direction $v \in \mathbb{R}^2$ from the depth image $o$. For the physical execution, we transform $v$ to a 3-D vector $u=(u_x,u_y,u_z)$ where the gripper pulls in the $x$-$y$ plane along $(u_x,u_y)$ while slightly lift along the $z$ axis about $u_z$. The pulling action ends before the gripper collides with the bin walls. The robot also performs a wiggling motion during pulling to reduce the effects of friction with the bin plane.
\end{itemize}

    The goal is to learn a policy $\pi_\Phi$ that maps the input depth image $o$ to the action $a\in \{a_\text{pick},a_\text{drop},a_\text{pull}\}$ where the trained networks PickNet and PullNet are parameterized as $\Phi$: $a \leftarrow \pi_\Phi(o)$. 

\section{Method}

    To efficiently pick up tangled-prone objects, the robot prioritizes grasping isolated objects in the clutter. If the bin contains no such objects, we leverage a buffer bin to reduce the degree of entanglement and help to perform the disentangling motions. The overview of our system is shown in Fig. \ref{fig:overview}. We first use a neural network PickNet to detect the untangled objects in the main bin. If such objects exist, the robot grasps them and transports them to the goal bin. Otherwise, the robot drops the entangled objects in a buffer bin to separate them. Then, the robot uses PickNet again to examine the buffer bin. If the objects are not successfully separated, we use a neural network PullNet to perform a pulling action and transport the singulated objects to the goal bin. The buffer bin helps to create an environment with few collisions for pulling. This process proceeds in iterations. 

\subsection{PickNet: Learning to Pick or Separate} \label{subsec:learn-pick}
    
    We learn PickNet $f_\text{pick}$ to (1) classify if the bin contains untangled objects for picking or if the robot should perform separation motions (dropping for the main bin and pulling for the buffer bin) and (2) predict the pixel-wise grasp affordance for picking and dropping actions. Given a depth image $o \in \mathbb{R}^{512\times512\times3}$ with triplicated depth values across three channels, the output is two heatmaps $f_\text{pick}(o) \in \mathbb{R}^{512\times512\times2}$: PickMap and SepMap. PickMap predicts the pixel-wise possibilities of picking untangled objects while SepMap calculates the possibilities of containing entangled objects. To infer the action in the main bin, we select the heatmap with the highest value between PickMap and SepMap to perform either picking or dropping action. In this case, the grasp position $p$ is selected at the highest pixel on the corresponding heatmap. For the buffer bin, if the maximum pixel on the PickMap is higher than that on the SepMap, the robot picks the objects to the goal bin. Otherwise, the robot performs the pulling motion, as inferred by our proposed PullNet. We use a ResNet-50 backbone \cite{he2016deep} with U-Net \cite{ronneberger2015u} skip connections for PickNet pre-trained on ImageNet \cite{deng2009imagenet}. We use an MSE loss during training. 

\begin{figure}[t]
\vspace{0.2cm}
    \centering
    \includegraphics[width=\linewidth]{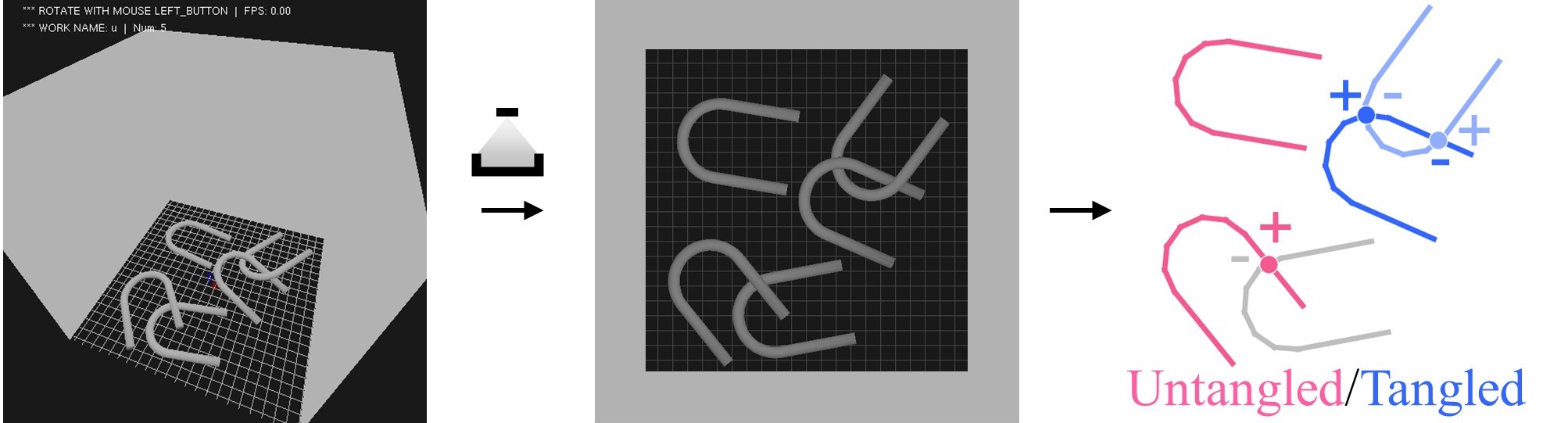}
    \caption{Process of distinguishing untangled/tangled objects in our algorithm. Given the full state of all objects as input, our algorithm skeletonizes the objects and obtains a graph collection by projecting along the vertical angle to the bin plane. For each object, we annotate the under-crossings it formed with others as $-1$ and otherwise as $+1$. The untangled objects (pink) are determined when the annotations of the crossings are $+1$ or without any crossings. The tangled objects (blue) have both $+1$ and $-1$ annotations. }
    \label{fig:sim-label}
\end{figure}

\subsection{PullNet: Learning to Pull for Separation} \label{subsec:learn-drag}

    We learn PullNet $f_\text{pull}$ to infer the pulling action including position $p$ and direction $v$. PullNet takes a depth image $o \in \mathbb{R}^{512\times512\times3}$ as input and generates a heatmap called PullMap $f_\text{pull}(o) \in \mathbb{R}^{512 \times 512}$ as output. Each pixel located in the PullMap represents the success possibility of pulling to the right of the image. We encode the pulling directions by rotating the input depth image for $\pi i/4, (i=0,1,\cdots,7)$ [rad]. PullNet can reason about pulling to the right for each rotated image. Then, the pulling direction $v$ and position $p$ are selected at the highest pixel value among eight PullMaps. We use a ResNet-18 \cite{he2016deep} backbone followed by a bi-linear upsampling layer pre-trained on ImageNet \cite{deng2009imagenet}. The network architecture is similar to  \cite{grannen2020untangling}. We use a binary cross-entropy loss for training. The pulling position is encoded as a 2D Gaussian. 

\subsection{Self-Supervised Data Generation}

    We develop a physics simulator using the NVIDIA PhysX library to collect synthetic data for PickNet and PullNet. We randomly drop 3D object models in a bin and use a simulated parallel gripper to execute consecutive pickings repeatedly. The picking process is executed under physical constraints. Instead of randomly exploring actions in the simulator, we propose an algorithmic supervisor that incorporates a set of entanglement representations, making it easier to control the collection process and adjust the dataset. Takes the full state of objects in the bin as input, our algorithm can (1) classify if the objects are tangled or not, (2) plan effective pulling actions for disentangling and (3) determine the sequence for picking demonstrations. 
    
    \textit{1) Algorithmic Supervisor:} To distinguish if the object is entangled, we leverage the method for skeletonization and crossing annotation in \cite{lui2013tangled} using the object models and poses. As Fig. \ref{fig:sim-label} shows, we first skeletonize each object into an undirected graph of nodes and edges. We project all objects onto the bin plane to obtain a collection of undirected graphs. We then calculate the crossings where the objects intersect with others and add them as nodes to the corresponding graph. We annotate the crossings that each object forms with others with $+1$ or $-1$. If the edge intersects above other objects, $+1$ is annotated for the corresponding object. Otherwise, $-1$ is annotated. From the graph collection using vertical projection, untangled objects have only $+1$ or no annotation while tangled objects have annotations of both $+1$ and $-1$. 
    
     Our algorithm first randomly drops the objects in the bin and selects untangled objects to grasp. If the bin contains no untangled objects but more than three entangled objects, the gripper picks the object with the least number of $-1$ annotations. If the bin only has less than three tangled objects, pulling is planned and performed. Note that our algorithm resets and drops the objects when the bin is empty or the gripper takes no objects out of the bin five times consecutively. 
     
\begin{figure}[t]
\vspace{0.2cm}
    \centering
    \includegraphics[width=\linewidth]{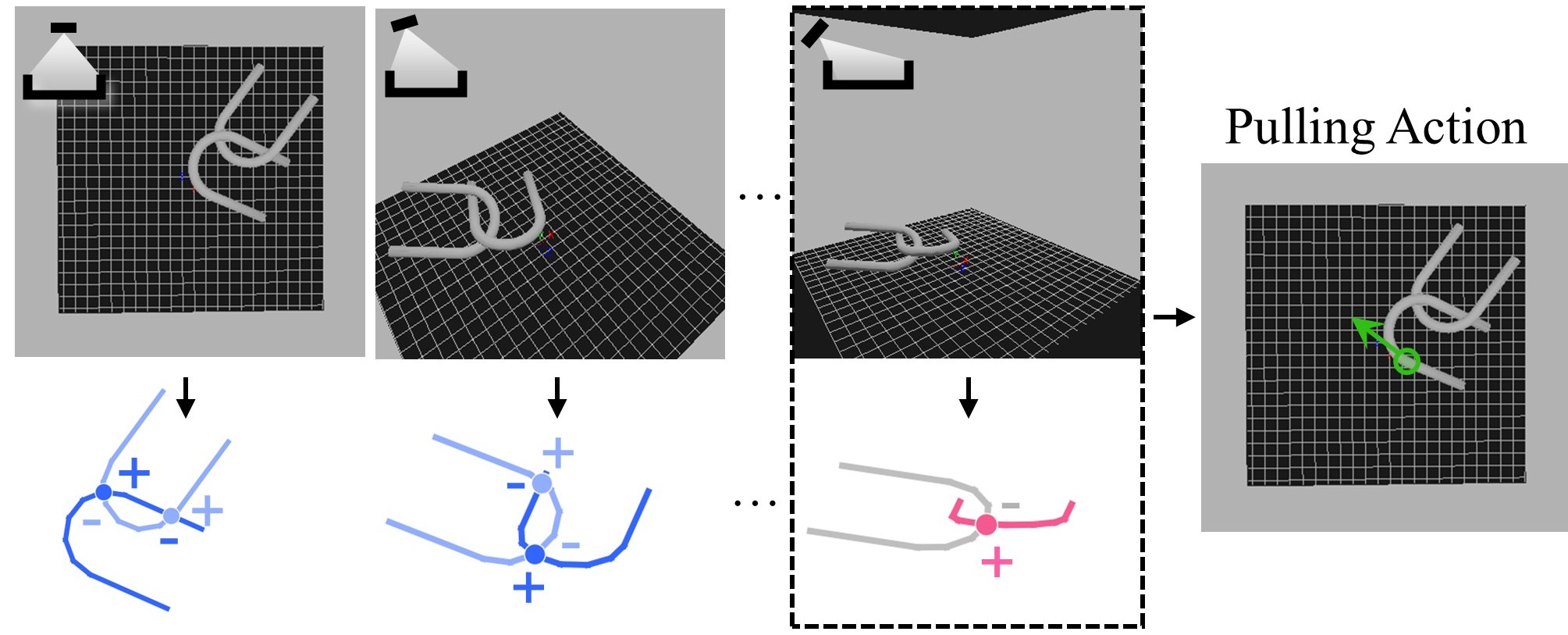}
    \caption{Process of calculating feasible directions and corresponding objects for pulling in our algorithm. By projecting and labeling the crossing from multiple angles, the feasible pulling direction is determined as the vector along the projection angle where the corresponding graph collection contains untangled (pink) objects.}
    \label{fig:sim-angle}
\end{figure}

     To plan pulling actions, we project objects from multiple angles to find feasible pulling directions (see Fig. \ref{fig:sim-angle}). If the graph collection of a projection angle contains untangled objects, it is possible to pulling this object out of entanglement along the corresponding projection angle. Thus, the feasible pulling direction $u$ is equivalent to the projection angle when the collection of projected graphs contains untangled objects. From a set of feasible pulling directions and pulling objects, our algorithm selects the object with the least number of $-1$ annotations in the vertical projection as the entanglement level of this object is expected to be lower than others. The grasp $(q,\theta)$ for pulling is selected by considering the non-collision grasps as in \cite{domae2014fast}. Specifically, we uniformly sample 48 projection angles as candidates in SO(3) space and define that the sampled angles should be in the range from $\pi/4$ [rad] to $\pi/2$ [rad] about the vertical axis to reduce the search cost.

    
\begin{figure}[t]
    \centering
    \includegraphics[width=\linewidth]{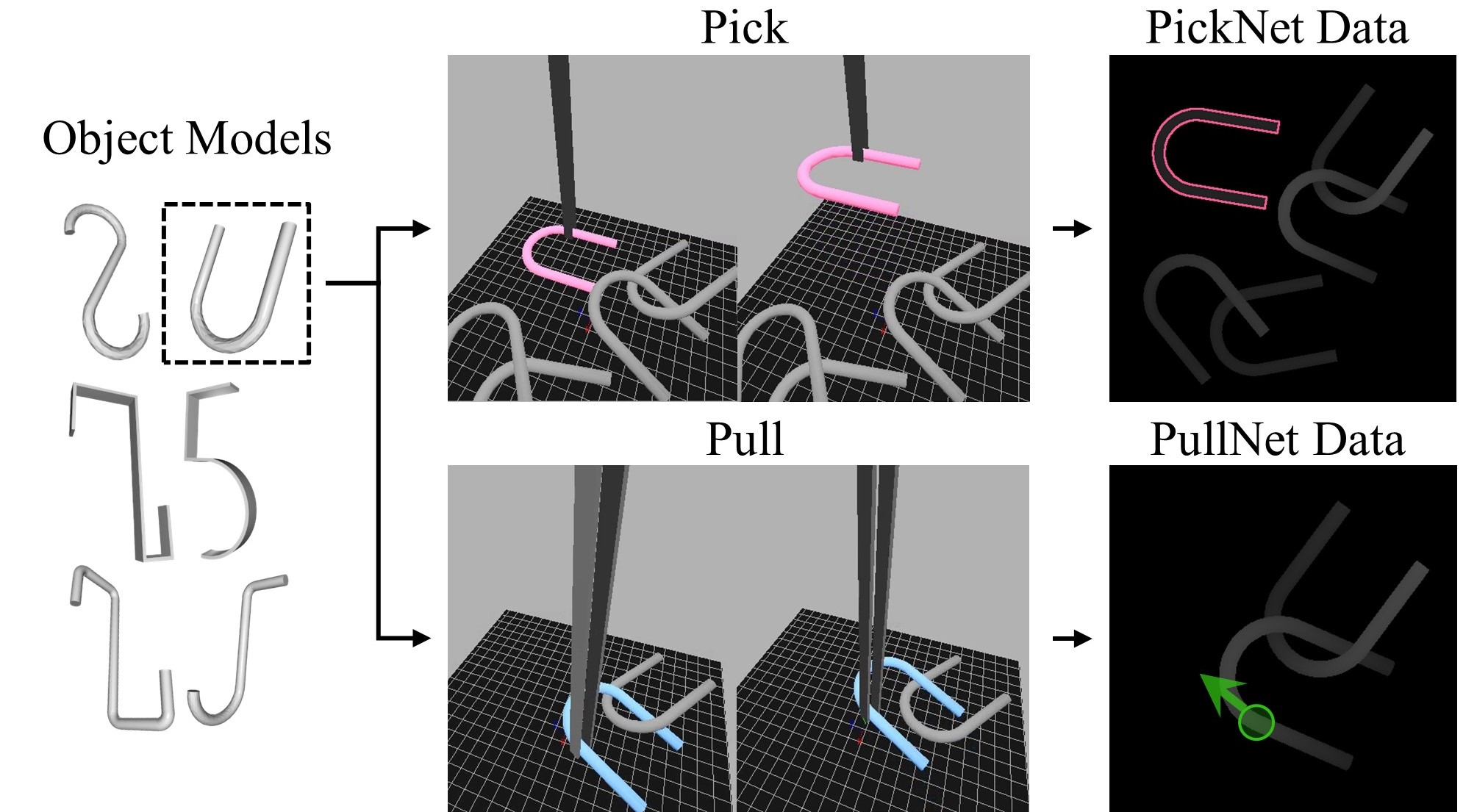}
    \caption{Demonstrations in simulation and data examples. }
    \label{fig:data-gen}
\end{figure}

    \textit{2) Training Datasets:} We use six models of tangled-prone objects including four planar objects and two non-planar objects.  Each sample for PickNet contains a depth image and two masks PickMap and SepMap. The process is shown in Fig. \ref{fig:data-gen}. After each attempt, if only one object is lifted without pulling, PickMap is masked with the target shape while SepMap is set to all zeros. Otherwise, if multiple objects are lifted, SepMap is masked with the target shape and PickMap is set to all zeros. On the other hand, if the gripper pulls and lifts only one object, the depth image and the pulling action are recorded to train PullNet. Each sample for PullNet contains a depth image and a Gaussian 2D encoding of the pulling point the same size as the depth image. The depth image is rotated so that the pulling direction points to the right in the image. After carrying out data augmentation methods, PickNet dataset contains 85,921 samples and PullNet dataset has 22,208 samples.
    
\section{Experiments and Results}\label{sec:exp}

\subsection{Experimental Setup}

    We use a NEXTAGE robot from Kawada Industries Inc. It operates over a workspace captured as a top-down depth image by a Photoneo PhoXi 3D scanner M. A parallel jaw gripper is attached at the tip of the left arm. In physical experiments, we use a PC with an Intel Core i7-CPU and 16GB memory with an Nvidia GeForce 1080 GPU. We use three seen objects and three unseen objects including a non-planar object for testing. When physically implementing the pulling action, we add a wiggling motion during pulling by rotating the wrist joint eight times by 0.1 [rad] with a velocity of 0.35 [rad/s]. The robot stops pulling before the gripper collides with the bin. When transforming the detected 2D pulling vector $v$ to the execution 3D vector $u=(u_x,u_y,u_z)$, we fix $u_z=0.2$ [cm]. 

    We compare the performance using two baselines and two versions of our policy. \textbf{FGE} is a model-free grasp detection algorithm using a depth image \cite{domae2014fast}. \textbf{EMap} takes a depth image as input and produces the entanglement map evaluating where contains entangled objects using edge information \cite{zhang2021topological}. \textbf{PickNet} is used in simulation to evaluate the ability to seek untangled objects. \textbf{PD} uses PickNet to detect graspable objects in the main bin and buffer bin. The tangled objects are transported to the buffer bin for separation. The robot performs dropping a maximum of three times and transports the objects to the goal bin. \textbf{PDP} denotes our complete workflow with both PickNet and PullNet and all three motion primitives. 

    We define four metrics to evaluate the performance of bin picking. Firstly,  let ``\# Goal attempts'' denote the times the robot transports one or multiple objects into the goal bin ($\text{\# } a_\text{pick} + \text{\# } a_\text{pull}$), ``\# Success attempts'' denote the times the robot transports only one object into the goal bin. ``\# Total attempts'' means the total times of executing all actions ($\text{\# } a_\text{pick}+ \text{\# } a_\text{drop}+\text{\# } a_\text{pull}$). \textbf{Success rate} ($\frac{\text{\# Success attempts}}{\text{\# Goal attempts}}$) evaluates the ability to grasp and transport a single object. \textbf{Completion} ($\frac{\text{\# Success attempts}}{\text{\# Objects}}$) evaluates the ability to accomplish the task of emptying the bin by picking up objects individually. \textbf{Action efficiency} ($\frac{\text{\# Success attempts}}{\text{\# Total attempts}}$) evaluates the effectiveness of our policy in utilizing picking, pulling and dropping actions to complete the task. \textbf{Mean Picks Per Hour (MPPH)} evaluates the computation and execution speed of the system.

\subsection{Simulated Experiments}
    
    In the simulation, we conduct a bin-picking task to evaluate the ability to seek untangled objects using FGE, EMap and PickNet. The methods detect the grasp position and feed to our simulator. The simulator locates the corresponding target object and automatically lifts it without gripper, excluding the irregular simulated physics phenomena in grasping or dynamical actions. The bin contains 30 objects and is replenished with the same number of objects after each attempt. We run 50 picking attempts for each object and evaluate the performance using the success rate only. 
    
    
    Table \ref{tab:res-picksim} shows the results of the simulated experiments. PickNet outperforms both baseline methods in success rates. FGE struggles with success rates as it can not discriminate if the target is entangled. EMap also becomes inefficient in dense clutter. Our policy significantly improves the success rates since the learned affordance map can seek untangled objects for such heavy occlusion. For unseen objects, our policy shows improvement in success rates compared to the baselines. The success rates of unseen objects on all methods are lower than seen objects, which could be attributed to the increased difficulty of the unseen set. The overall success rates are not impressive since when the bin contains no isolated objects, there is no separation motion to perform. It demonstrates the necessity of separation strategies to handle unsolvable cases relying solely on PickNet.


\begin{table}[t]
\vspace{0.2cm}
\footnotesize
\renewcommand\arraystretch{1}
    \caption{Results of Simulated Experiments}
    \centering
    \setlength\tabcolsep{4.4pt} 
    \begin{tabular}{c|cccc|cccc}
        \toprule
        \vspace{2pt}
        & \multicolumn{4}{c|}{Seen} & \multicolumn{4}{c}{Unseen} \\
        &\includegraphics[height=30px]{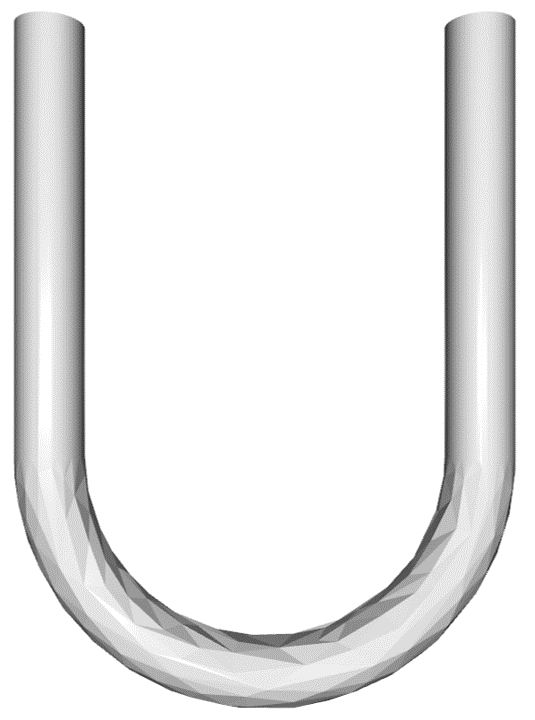} & \includegraphics[height=30px]{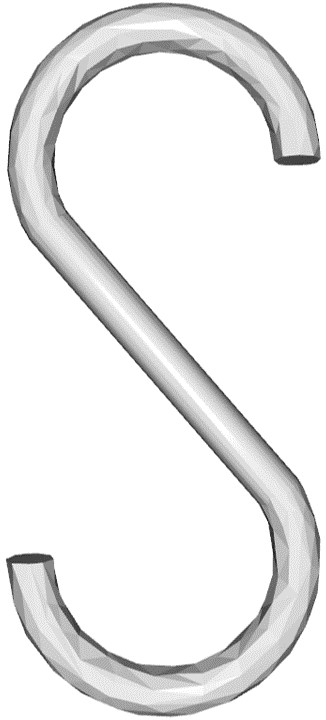} & \includegraphics[height=30px]{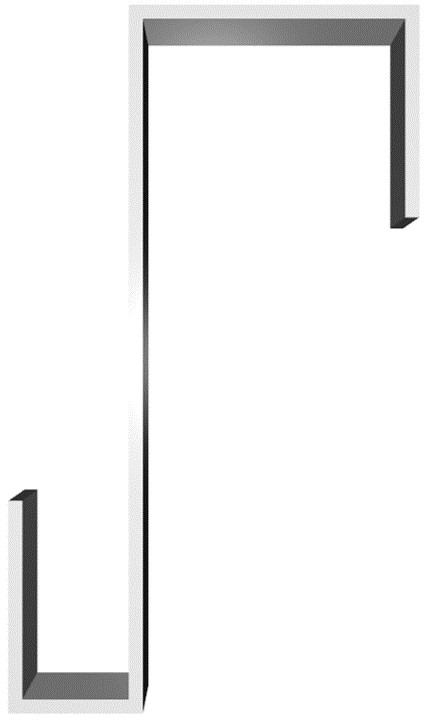} & Avg.
        &\includegraphics[height=30px]{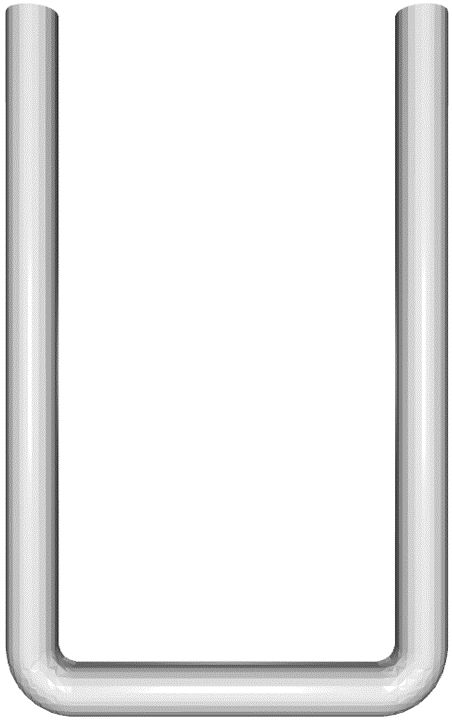} & \includegraphics[height=30px]{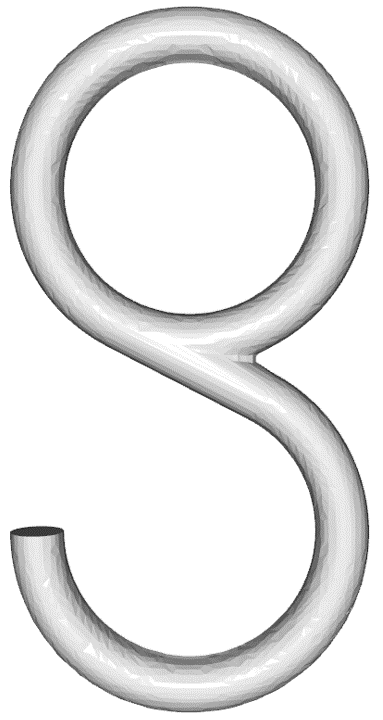} & \includegraphics[height=30px]{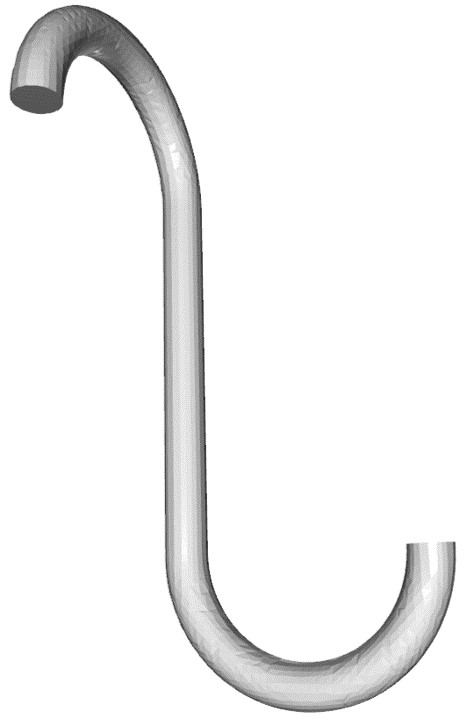} & Avg. \\
        \midrule
        & \multicolumn{8}{c}{Success Rate (\%)}\\
        \midrule
        FGE    & 60.0 & 50.0 & 44.0 & 51.3 & 30.0 & 60.0 & 46.0 & 48.7 \\
        EMap   & 58.0 & 52.0 & 54.0 & 54.7 & 26.0 & 52.0 & 42.0 & 40.0 \\
        PickNet & 92.0 & 86.0 & 82.0 & \textbf{86.7} & 60.0 & 88.0 & 57.0 &  \textbf{68.3} \\
        \bottomrule
    \end{tabular}
    \label{tab:res-picksim}
\end{table}

\subsection{Real-World Experiments}

    Different from the simulated experiments where the robot is required to pick from a bin containing 30 objects every time, the goal of the real-world task is to empty the bin filled with 20 objects. We run three tests for each object using each method.  
   
    \textit{1) Comparison with Baselines:} The results of bin picking in the real world are shown in Table \ref{tab:res-empty}. PD and PDP outperform baselines FGE and EMap in all metrics. Our policies can perform the task with a success rate of around 90\%, almost as high as that of picking simple-shaped objects. Compared with FGE, our policies can detect potentially entangled objects. The affordance map learned by PickNet can also indicate the state of the entanglement more explicitly than EMap. The proposed separation strategies are useful to improve the performance when picking such entangled objects. The completion results suggest that our methods outperform the baselines, demonstrating their ability to complete the task of emptying the bin. Action efficiency suggests our policy PDP performs the best among all methods including PD. PD requires extra actions to separate the entangled objects from the buffer bin while PDP can disentangle and transport by only one action using PullNet. Finally, we compare the speed of each system using the metric MPPH. Our policy can achieve more than 200 mean picks per hour. Specifically, PDP with both networks achieves the highest MPPH than PD since PD requires more actions to complete the task. 

\begin{table}[t]
\vspace{0.2cm}
\caption{Results of Real-World Experiments}
\footnotesize
\renewcommand\arraystretch{1}
    \centering
    \setlength\tabcolsep{4.5pt} 
    \begin{tabular}{c|cccc|cccc}
        \toprule
        \vspace{2pt}
        & \multicolumn{4}{c|}{Seen} & \multicolumn{4}{c}{Unseen} \\
        &\includegraphics[height=30px]{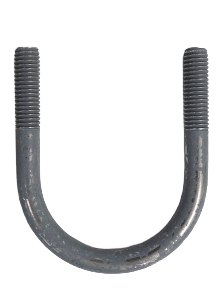} & \includegraphics[height=30px]{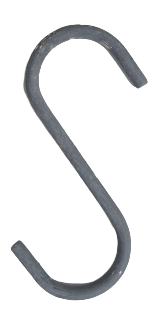} & \includegraphics[height=30px]{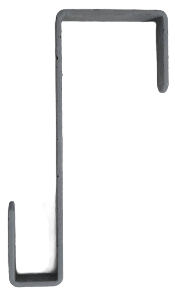} & Avg.  
        &\includegraphics[height=30px]{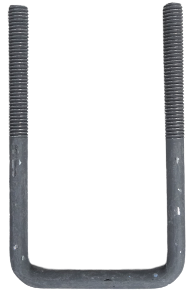} & \includegraphics[height=30px]{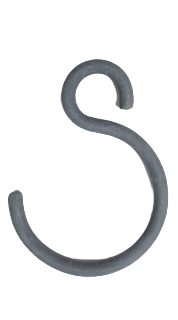} & \includegraphics[height=30px]{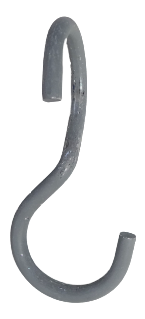} & Avg. \\
        \midrule
        \multicolumn{9}{c}{Success Rate (\%)} \\
        \midrule
        FGE    & 67.5 & 60.0 & 67.6 & 65.0 & 67.4 & 68.4 & 42.3 & 59.0 \\
        EMap   & 71.4 & 61.5 & 73.0 & 70.6 & 74.9 & 68.9 & 42.3 & 62.0 \\
        PD & 95.2 & 85.3 & 86.9 & 89.2 & 89.2 & 89.2 & 83.3 & 87.2 \\
        PDP  & 95.1 & 85.2 & 91.7 & \textbf{90.7} & 88.2 & 93.3 & 84.5 & \textbf{88.7} \\
        \midrule
        \midrule
        \multicolumn{9}{c}{Completion (\%)} \\
        \midrule
        FGE    & 54.0 & 30.0 & 38.3 & 40.8 & 48.3 & 45.0 & 41.7 & 44.4 \\
        EMap   & 50.0 & 45.0 & 45.0 & 47.0 & 58.3 & 50.0 & 41.7 & 50.0 \\
        PD & 96.7 & 78.3 & 88.3 & 87.8 & 86.7 & 93.3 & 76.7 & \textbf{85.6} \\
        PDP  & 96.7 & 78.3 & 93.3 & \textbf{89.4} & 86.7 & 93.3 & 73.3 & 84.4 \\
        \midrule
        \midrule
        \multicolumn{9}{c}{Action Efficiency (\%)} \\
        \midrule
        FGE    & 67.5 & 60.0 & 67.6 & 65.0 & 67.4 & 68.4 & 42.3 & 59.0 \\
        EMap   & 71.4 & 67.5 & 73.0 & 70.6 & 74.9 & 68.9 & 42.3 & 62.0 \\
        PD & 73.4 & 64.2 & 73.6 & 70.4 & 70.1 & 63.8 & 61.8 & 65.2 \\
        PDP  & 84.0 & 72.2 & 75.8 & \textbf{77.3} & 73.2 & 72.0 & 68.2 & \textbf{71.1} \\
        \midrule
        \multicolumn{9}{c}{Mean Picks Per Hour (MPPH)}\\
        \midrule
        FGE & \multicolumn{8}{c}{171}\\
        EMap & \multicolumn{8}{c}{150}\\
        PD & \multicolumn{8}{c}{203}\\
        PDP & \multicolumn{8}{c}{\textbf{220}}\\
        \bottomrule
    \end{tabular}
    \label{tab:res-empty}
\end{table}

\begin{figure*}[t]
\vspace{0.2cm}
    \centering
    \includegraphics[width=\linewidth]{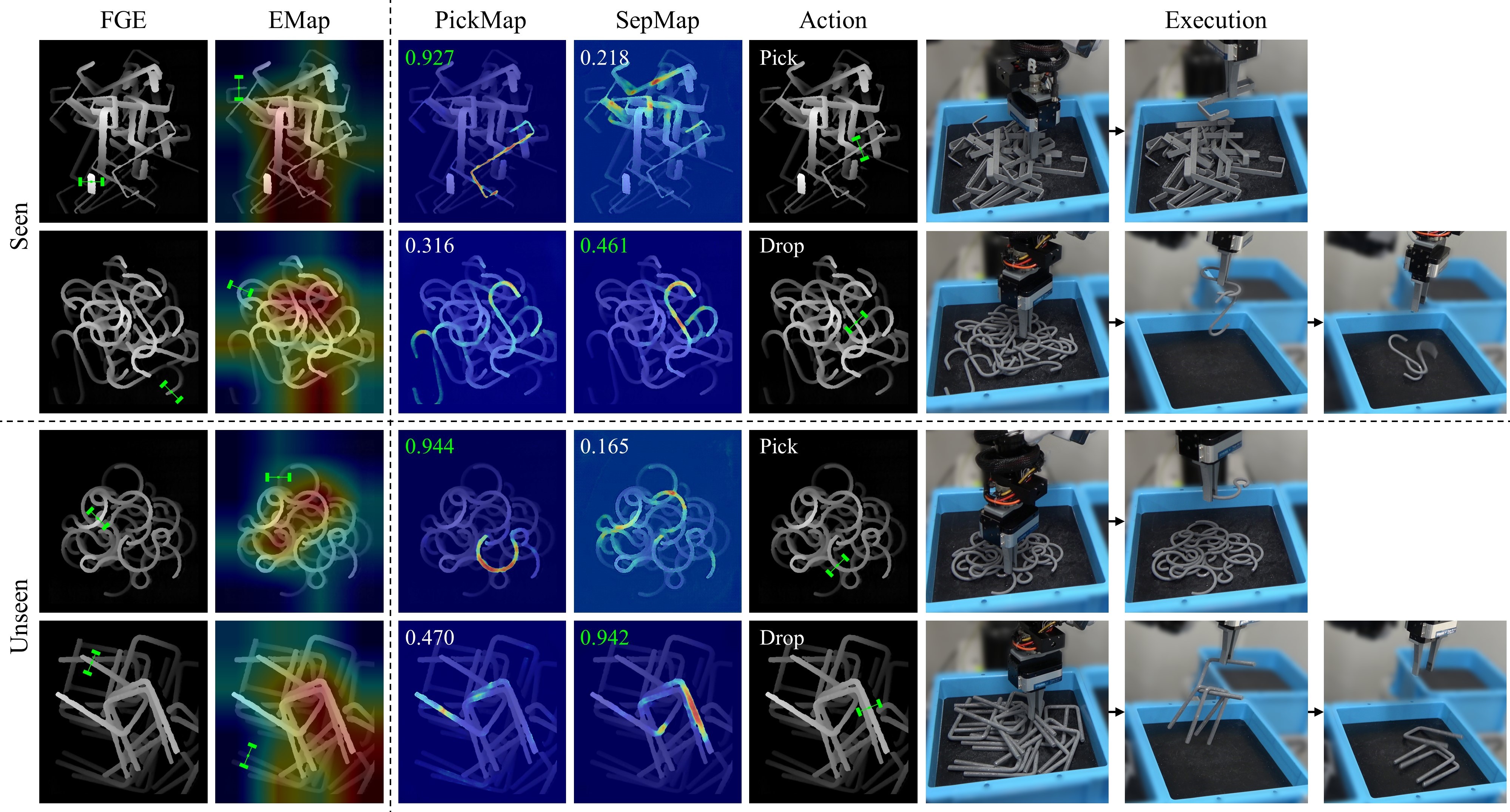}
    \caption{Qualitative results using PickNet and the corresponding physical executions. Using the same depth map as input, we also present the detected grasps using FGE, the grasps and the entanglement maps using EMap (red regions show high possibilities of containing entangled objects). PickNet outputs PickMap and SepMap with their maximum pixel value as the affordance of picking or dropping. }
    \label{fig:ret-pn}
\end{figure*}

\begin{figure*}[t]
    \centering
    \includegraphics[width=\linewidth]{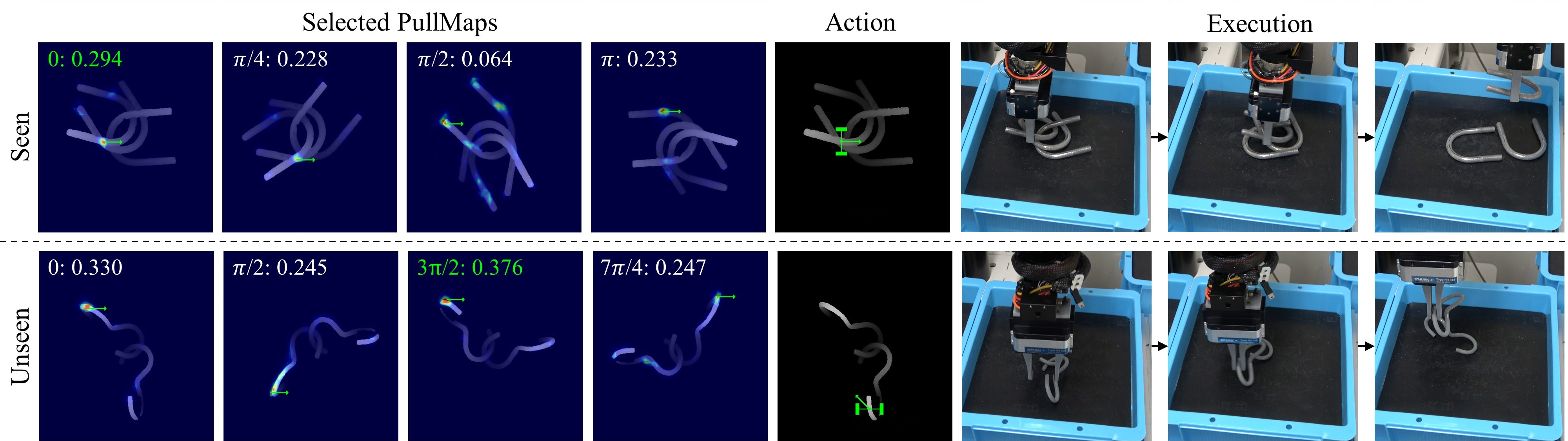}
    \caption{Qualitative results of PullNet and the corresponding physical executions. PullNet predicts the position and direction for pulling. We rotate the input depth image in eight directions and present four selected PullMaps with their maximum pixel values. The action is selected by the highest score among all PullMaps. The green arrows denote the pulling directions. }
    \label{fig:ret-sn}
\end{figure*}

    \textit{2) Does Dropping Help? } We investigate the efficiency of the dropping motion. As Table \ref{tab:res-empty} shows, PD uses dropping action as the only separation strategy and achieves a similar success rate and task completion as our complete policy PDP for both seen and unseen objects. The dropping actions can (1) effectively disentangle the grasped objects and (2) reduce the degree of entanglement, thereby creating space for subsequent disentangling manipulation. Both PD and PDP benefit from this motion. However, the action efficiency of PD is significantly lower than that of PDP. Unlike pulling, dropping acts as an intermediate action and cannot separate and transport the objects to goal bin simultaneously. Table \ref{tab:action-rate} shows the number of dropping actions in real-world experiments. PD requires more dropping actions than PDP. In cases where the objects are still entangled after the first time of dropping in the buffer bin, PD repeatedly performs this action until the objects are singulated or it reaches three times. We also observe that dropping cannot solve some difficult entanglement cases without well-planned motions or modeling of physical effects. For these reasons, only relying dropping as the separation strategy results in lower action efficiency. 
    
    \textit{3) Does Pulling Help? } We can observe that PDP, incorporating both separation strategies of dropping and pulling, achieves the best performance particularly in terms of action efficiency and MPPH. Unlike dropping actions, pulling requires motion planning based on visual observation, making it more interpretable for skillful tasks such as object disentangling. As Table \ref{tab:res-empty} shows, the success rates are relatively higher when using pulling compared to dropping, as pulling actions can resolve challenging entanglement cases. Pulling also contributes to the action efficiency. PDP follows a hierarchical strategy by first dropping the objects to create a relatively simple entanglement state and then planning pulling actions to further disentangle them. With the reduced degrees of entanglement by dropping in the buffer bin, pulling can efficiently disentangle and transport. Table \ref{tab:action-rate} shows the successful pulling rates among all actions. Pulling and dropping in our policy PDP can be orchestrated together to achieve the best performance for picking tangled-prone objects. 
    
\begin{table}[t]
\vspace{0.2cm}
\caption{Distribution of Actions in Real-World Experiments}
\footnotesize
\renewcommand\arraystretch{1}
    \centering
    \begin{tabular}{c|ccc|ccc}
    \toprule
    \vspace{2pt}
        &\multicolumn{3}{c|}{Seen} & \multicolumn{3}{c}{Unseen} \\
        &\includegraphics[height=30px]{img/u.png} & \includegraphics[height=30px]{img/sc.png} & \includegraphics[height=30px]{img/sr.png} 
        &\includegraphics[height=30px]{img/ul.png} & \includegraphics[height=30px]{img/se.png} & \includegraphics[height=30px]{img/sn.png}\\
        \midrule
        \multicolumn{7}{c}{Dropping Rate (\%)}\\
        \midrule
        PD & 19.0 & 25.7 & 24.7 & 21.6 & 26.1 & 28.4 \\
        PDP  & 10.1 & 12.3 & 16.2 & 15.5 & 17.9 & 15.4 \\
        \midrule
        \multicolumn{7}{c}{Pulling Rate (\%)}\\
        \midrule
        PDP  & 1.45 & 4.62 & 2.70 & 1.41 & 2.56 & 7.69 \\
        \midrule
        \multicolumn{7}{c}{Successful Pulling Rate (\%)}\\
        \midrule
        PDP  & 1.45 & 3.08 & 2.70 & 1.41 & 2.56 & 6.15 \\
        \bottomrule
    \end{tabular}
    \label{tab:action-rate}
\end{table}

    \textit{4) Generalization to Unseen Objects:} Finally, we evaluate the performance of our policies using unseen objects. Table \ref{tab:res-empty} demonstrates that our policies can be generalized to novel objects. Both PD and PDP can recognize isolated objects even if for the challenging non-planar object (the last column in Table \ref{tab:res-empty}). Thanks to our efficient data collection algorithm, which allows us to collect a large-scale of synthetic data, our networks are capable of handling unknown object geometries and various entanglement scenarios. However, all metrics for unseen objects are slightly lower than seen objects in both PD and PDP. We can assume by the performance of the model-free method FGE that unseen objects pose additional challenges, such as heavy occlusion of non-planar objects. PickNet may occasionally misidentify isolated objects as entangled objects and predicts redundant separation actions. Additionally, we present the visualized results using PickNet and PullNet in Fig. \ref{fig:ret-pn} in Fig. \ref{fig:ret-sn}. We also visualize the results from FGE and EMap using the same observation as our policy. These visualizations demonstrate that our policy can accurately extract geometrical information for isolated objects and better represent entanglement compared to the baselines. 

\subsection{Failure Modes and Limitations}
    We observe some failure modes during physical experiments and investigate the limitations of our method. We divide them into three categories as follows. 

    \textit{1) Grasp Failure:} The average grasp failure of our policy (PD and PDP) is 4.8\%. Grasp fails when there is no collision-free orientation around the predicted grasp point, making the gripper collide with the objects or the environment. We also observe some common failure modes of bin picking, such as objects positioned against walls, which left no space for grasping. 
    
    \textit{2) Challenging Entanglement Patterns: } The proposed PickNet and PullNet have limitations. First, when the target object forms an endless chain with others, the robot cannot entirely lift and drop the whole chain in the buffer bin. It is also difficult to visually predict how many objects will be grasped based on a top-down depth map. On the other hand, the proposed separation strategies (dropping and pulling) cannot handle several entanglement cases where the objects are tightly wedged together, requiring multi-step or bimanual manipulation to solve. 

    \textit{3) Unsuitable Object Shapes:} Some object shapes are unsuitable for our policy, such as a tree-like shape since our dataset only includes linear shapes. In the future, we will extend our policy to incorporate objects with various shapes. It will be interesting to collect data only using a minimal amount of objects based on the entanglement representation of their geometries. 
    


\section{Conclusion}\label{sec:con}
    
    We propose a bin-picking system for efficiently picking tangled-prone objects. Our hierarchy method is learned using self-supervised simulated data, enabling the robot to perform picking or separation actions dexterously based on the visual observations. Experimental results show the effectiveness of the proposed separation strategies. Our policy outperforms baseline methods in completing the challenging task of emptying the bin with tangled-prone objects with higher success rates and efficiency. We further demonstrate the generalization of our policy using novel objects. In the future, we will extend our policies by leveraging various sensing or more skillful motion primitives for more complex-shaped or deformable objects. 

\bibliographystyle{ieeetr}
\bibliography{ebibsample.bib}

\end{document}


\makeatletter
\let\oldtwocolumn\twocolumn
\renewcommand\twocolumn[1][]{%
    \oldtwocolumn[{#1}{
    \begin{center}
           \includegraphics[width=\textwidth]{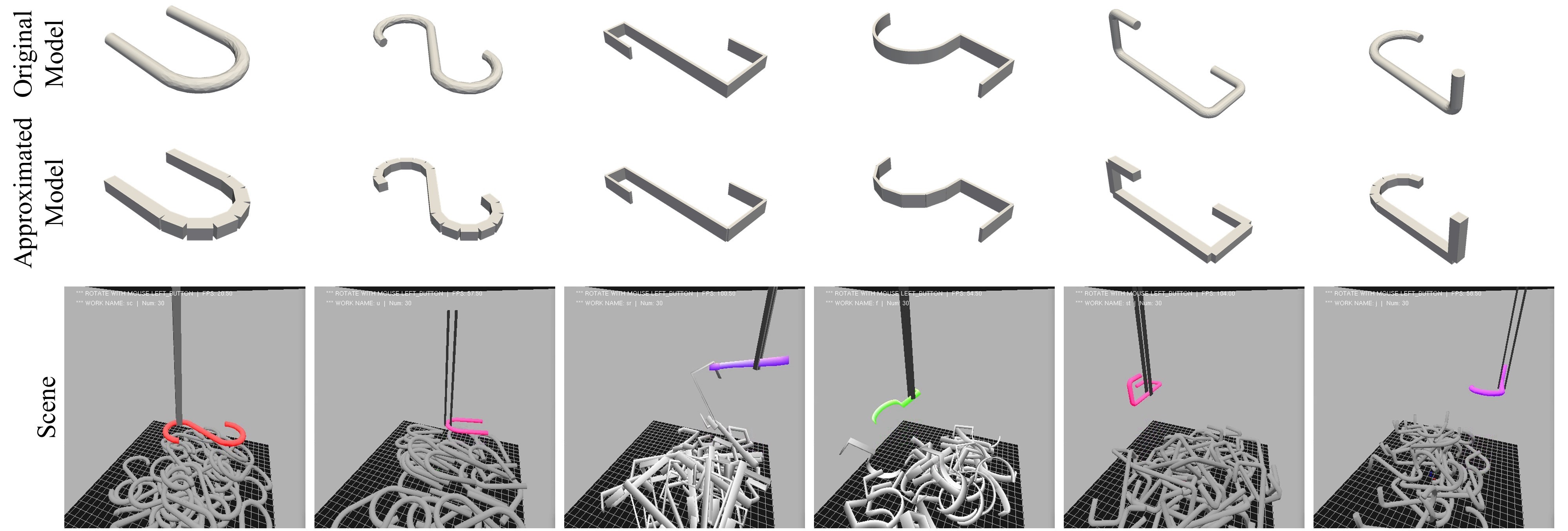}
           \captionof{figure}{Objects and scenes.}
           \label{fig:sim-approx-obj}
        \end{center}
    }]
}
\makeatother

\maketitle
\thispagestyle{plain}
\pagestyle{plain}

    The supplementary material is structured as follows: Section S.I contains details of the physics simulator and data collection. Section S.II specifies details on training PickNet and PullNet. Section S.III contains additional implementation details on the grasp orientation detection algorithm or other visualized results.

\section{Physics Simulator Details}
\subsection{Physics Simulator}
    We use NVIDIA’s PhysX physics engine to collect synthetic data. We approximate the objects as a set of rigid-body cuboids to (1) balance the trade-off between the simulation accuracy and calculation time and (2) decrease the effects of unreal physical phenomenon when computing collisions in clutter. We model the parallel jaw gripper as two parallel cuboids and the bin as five rigid-body planes. We manually adjust the size and physical parameters of the rigid bodies to achieve the similar interaction behaviors as that of the real-world. Table \ref{tab:sim-param} shows these parameters. We also present the origin model, the approximated model, the clutter scene and the moment of grasping of each object used in the data generation process in Fig \ref{fig:sim-approx-obj}. Since we only use depth maps as dataset, we do not consider the visual appearance of the objects such as textures. Moreover, the shapes of the objects are determined and designed based on the previous works of bin picking.  

\begin{table}[t]
\renewcommand\arraystretch{1}
\centering
\small
\caption{Physics Simulator Parameters}
    \begin{tabular}{@{\extracolsep}ll}
    \toprule
    Parameters & Value\\
    \midrule
    Bin Static Friction & 0.40\\
    Bin Dynamic Friction & 0.35\\
    Bin coefficient of restitution & 0.05\\
    Bin Size & (22.5,22.5,22.5) cm\\ 
    Object Static Friction & 0.30\\
    Object Dynamic Friction & 0.25\\
    Object coefficient of restitution & 0.40 \\
    Object Density & 1 g/cm$\rm ^3$\\
    \bottomrule
    \end{tabular}
\label{tab:sim-param}
\end{table}


\begin{algorithm}[t]
\small
\SetAlgoLined
\SetKwFunction{AC}{AnnotateCrossing}
\SetKwFunction{DG}{DetectGrasp}
\SetKwFunction{RT}{RecogTangle}
\SetKwFunction{PP}{PlanPulling}
    \SetKwProg{Fn}{Function}{:}{}
    \Fn{\RT{}}
    {
        $\{G_0,G_1,...\} \gets$ vertically projected objects\;
        \For{$G_i,i \gets 1,2,...$}
            {$X(G_i) \gets $ \AC{}\;}
            \uIf{$X(G_i)$ \rm is empty}
                {$q,\theta \gets$ \DG{$i$}\;
                \textbf{return} $a_\text{pick}=(q,\theta$)\;}
            \uElseIf{$\forall x \in X(G_i)=+1$}
                {$q,\theta \gets$ \DG{$i$}\;
                \textbf{return} $a_\text{pick}=(q,\theta$)\;}
            \uElseIf{\rm bin contains less than three objects}
                {$i,u \gets$ \PP{}\;
                \textbf{return} $a_\text{pull}=(q,\theta,u)$\;}
            \uElse
                {$i \gets X(G_i)$ with minimal number of $-1$ \;
                $q,\theta \gets$ \DG{$i$}\;
                \textbf{return} $a_\text{pick}=(q,\theta$)\;}
    }
\caption{Tangle Recognition Function}
\label{alg:supervisor-tangle}
\end{algorithm}

    Meanwhile, we explain how the grasp is executed in the simulator under physical constraints. The policy includes three parameters: 
\begin{itemize}
    \item $v_g^\text{close}$: Velocity of moving the fingers for closing. 
    \item $v_g^\text{lift}$: Velocity of the fingers for lifting. 
    \item $d_g$: Distance between two fingers. 
\end{itemize}
    First, the gripper approaches the target object using a 3D position and an orientation angle calculated by our grasp detection algorithm. Then, to let the gripper contact with the object, we set a closing speed $v_g^\text{close}$ acted as grasping force. If $v_g^\text{close} \rightarrow 0,d_g>0$, the target is grasped. Next, the gripper lifts with the grasped object by a fixed lifting speed $v_g^\text{lift}$ and an adjusted closing speed $v_g^\text{close}$. $v_g^\text{close}$ is calculated based on the force where two fingers act on the object. During lifting, if $d_g=0$, which means the object is slipped from the gripper, $v_g^\text{lift}$ remains the same while $v_g^\text{close}=0$. Finally, the grasping process is terminated if the gripper is outside the bin. 

    We also controls the pulling process similar as the grasping or lifting process. All actions are perform in the simulated physical environment.

\subsection{Algorithmic Supervisor}

    We implement an algorithm to collect data in a self-supervised manner. Our algorithmic supervisor has three functions as follows. 

    \textit{1) Tangle Recognition:} . Algorithm \ref{alg:supervisor-tangle} shows the detail of \texttt{RecogTangle()}. First, we skeletonize each object into an undirected graph consisting of nodes and edges. We project all objects onto the bin plane to obtain a collection of undirected graphs $G_0,G_1,...$. We then compute and annotate each crossings where each object intersect with others. Each object has an annotation list $X(G_i)$: a collection of $+1$ and $-1$ where $i$ denotes the index of the object. If the edge intersects above the edge of other objects, $+1$ is annotated for the corresponding object. Otherwise, $-1$ is annotated. From the graph collection using vertical projection, untangled objects have only $+1$ or no annotation while tangled objects have annotations of both $+1$ and $-1$. 
    
    \texttt{RecogTangle()} finally returns an action $a$ under four conditions: (1) If there exists an empty annotation list, the gripper lift the corresponding object; (2) Otherwise, ff there exists an annotation list where all elements equal $+1$, the gripper lift the corresponding object; (3) Otherwise, it means that the bin only contains the entangled objects, if the bin contains less than three objects, we leverage \texttt{PlanPulling()} to disentangling them; (4) Otherwise, if the bin contains more than three entangled objects, the gripper lift the one with the least number of $-1$. Finally, we detect the grasp using the depth image and the mask of the target object using \texttt{DetectGrasp()}. The details of this function are elaborated on Section \ref{subsec:exp-grasp} in this supplementary materials. 

\begin{algorithm}[t]
\small
\SetAlgoLined
\SetKwFunction{PP}{PlanPulling}
\SetKwFunction{Select}{Select}
    \SetKwProg{Fn}{Function}{:}{}
    \Fn{\PP{}}
    {
        $u_0,u_1,... \gets$ sampled directions for pulling\;
        $S \gets$ empty list\;
        \For{$u_j \gets 0,1,...$}
            {
            $\{G^\prime_0,G^\prime_1,...\} \gets$ projected objects along $u_j$\;
            \For{$G^\prime_i,i \gets 1,2,...$}
                {$X(G^\prime_i) \gets $ \AC{}\;}
            \uIf{$G^\prime_i$ \rm is empty \textbf{or} $\forall x \in X(G^\prime_i)=+1$}
                {Append $(i,u_j)$ to $S$\;}
            }
            \textbf{return} $i^*,u^* \gets$ \Select{$S$}\;  
    }
\caption{Pulling Planning Function}
\label{alg:supervisor-pull}
\end{algorithm}

\begin{figure}[t]
    \centering
    \includegraphics[width=\linewidth]{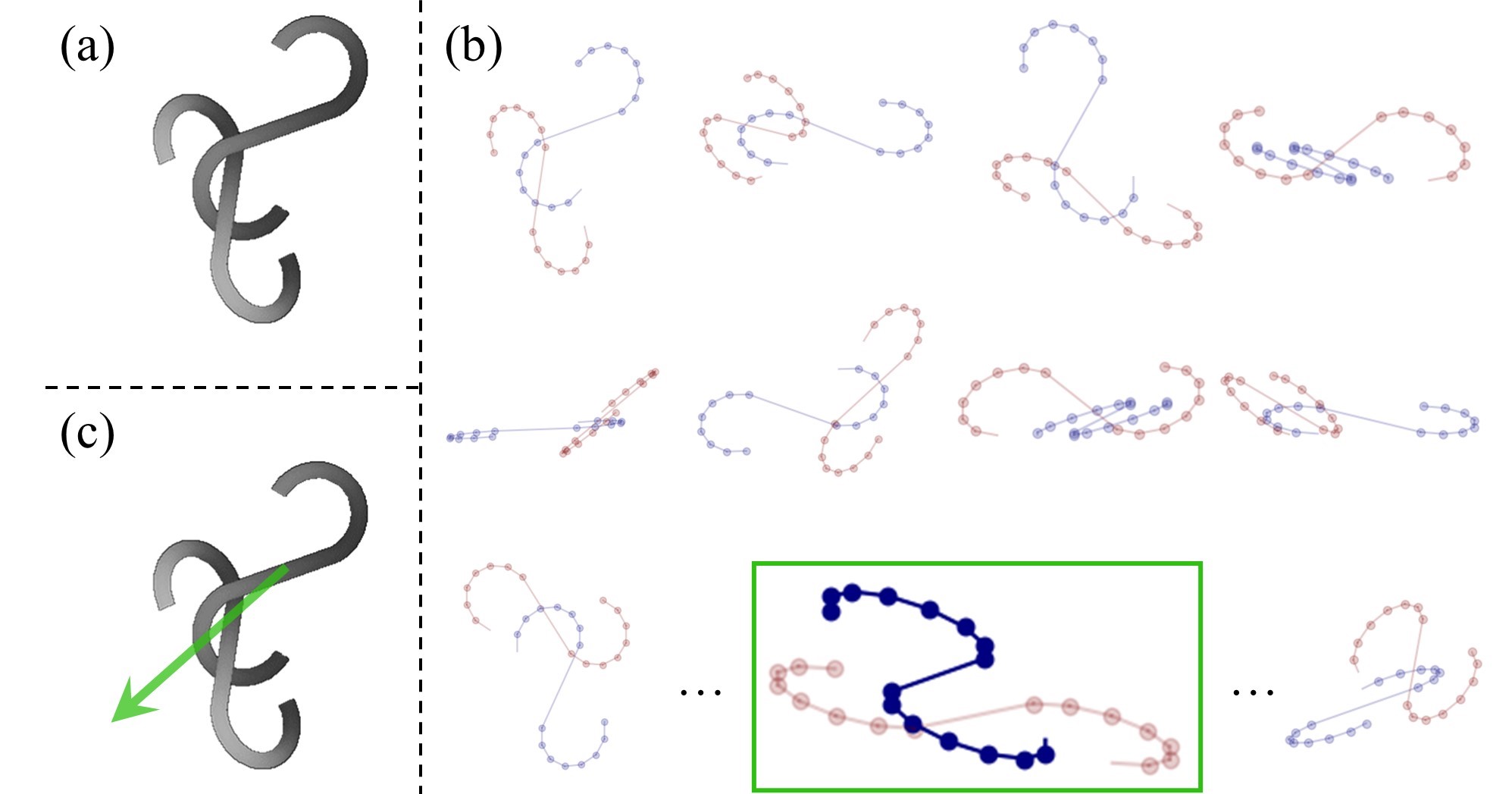}
    \caption{(a) Two entangled objects. (b) Some projected graphs using different sampled directions. Figure with green block refers to a solvable direction where the crossing annotation of the blue objects only contains $+1$. (c) Visualized pulling direction. }
    \label{fig:tangle-plan}
\end{figure}

    \textit{2) Pulling Planning:} As Algorithm \ref{alg:supervisor-pull} shows, we first sample a set of projection angles represented by 3-D vectors $u_0,u_1,...$. For each vector $u_j$, we project each object along $u_j$ to obtain a undirected graph collection $\{G^\prime_0,G^\prime_1,...\}$. Function \texttt{AnnotateCrossing()} is used to annotated crossings $X(G^\prime_i)$ for $G^\prime_i$. Next, we save the object $i$, projection direction $u_j$ where $X(G_i)$ is empty or exists only $+1$ labels as the pulling candidates. Fig. \ref{fig:tangle-plan} shows some projected graph collections. From the saved pulling candidates, we leverage some heuristics (\texttt{Select()} in line 12) to select the best pulling direction and objects. We check the annotations of each object in the vertically projected graph and compute the pulling distance along the corresponding pulling vector before each object hits the bin walls. We select the candidate where the objects has at least number of $-1$ annotations and the maximum pulling distance as the best pulling action $i^*,u^*$. The grasp $(q,\theta)$ is computed by the same function \texttt{DetectGrasp()}. 

    
    \textit{3) Picking Demonstration:} Algorithm \ref{alg:supervisor} shows the complete process of data collection during simulated demontrations. First, objects are randomly dropped in to the bin (line 2). Line 5 denotes the function \texttt{RecogTangle} of this algorithm, which returns the picking or pulling actions $a$ for the execution. After detecting the grasping object and executing the corresponding action (line 6), one attempt is terminated when the gripper is out of the bin. Then, we count the number of objects in the bin before and after the attempt. If only one object is taken out of the bin, we record the data including the depth image, mask and corresponding action (line 7-13). Otherwise, the count of failure attempts adds one and the simulator tries again to find the grasp and action (line 14-15). If the number of failed attempts exceeds five (line 16-17), the bin is reloaded by randomly dropping the objects (line 2) and resetting the number of failed attempts (line 3).

\begin{algorithm}[t]
\small
\SetAlgoLined
\SetKwInput{KwInput}{Input} 
\SetKwInput{KwOutput}{Output} 
\SetKwFunction{RT}{RecogTangle}
\SetKwFunction{PP}{PlanPulling}
\While{True} {
    Drop objects in the bin\;
    $N_\text{fail} \gets 0$\;
    \While{\rm bin contains objects}
    {   
        $a \gets$ \RT()\;
        Execute $a$\;
        \uIf{\rm only one object is out of the bin}
            {
            \uIf{\rm $a_\text{pull}$ is executed}
            {Record for PickNet (masked SepMap) and PullNet\;}
            \uElse{
            Record for PickNet (masked PickMap)\;}}
        \uElseIf{\rm more than one object is out of the bin}
            {Record for PickNet (masked SepMap)\;}
        \uElse
            {$N_\text{fail} \gets N_\text{fail} + 1$\;}
        \uIf{\rm $N_\text{fail} > 5$}
        {\textbf{Continue}\;}
    }
}
\caption{Algorithmic Supervisor}
\label{alg:supervisor}
\end{algorithm}

\section{Training Details}

\subsection{PickNet}

    \textit{1) Dataset:} The ground truth of a PickNet data sample is a 2-channel binary map, PickMap and SepMap shown in Fig. \ref{fig:data-gt}(a). During data collections, our algorithm first selects the untangled objects for picking. After successfully picking one object, we record the binary mask of the complete shape of this object and use it as PickMap while the SepMap is set to all zeros. Our algorithm continuous seeks untangled objects until the bin contains no such objects. Then, after the entangled object is grasped, we record its complete shape as SepMap while PickMap is set to all zeros. We augmented the datasets by image-based transformations as Table \ref{tab:data-aug} shows. We also provide some examples of the data augmentation in Fig. \ref{fig:data-aug}. Finally, we augmented the PickNet dataset 2X to 85,921 samples.


    \textit{2) Training details:} PickNet learns a mapping function $o \in \mathbb{R}^{512 \times 512 \times 3} \rightarrow f_\text{pick}(o) \in \mathbb{R}^{512 \times 512 \times 2}$. The input We triplicate depth values across three channels to match with the default input size of the pretrained backbone ResNet. we use a ResNet-50 pre-trained on Imagenet with U-Net skip connections to train PickNet. We use the mean square error (MSE) as loss function. We train PickNet with a batch size of 2 using the stochastic gradient descent (SGD) optimizer with a learning rate of 0.001 and a weight decay of 0.0001 on a Nvidia GeForce RTX 3080 GPU. We finally select the weights from the 8-th epoch since it achieve the best performance.

\subsection{PullNet}

\begin{figure}[t]
    \centering
    \includegraphics[width=\linewidth]{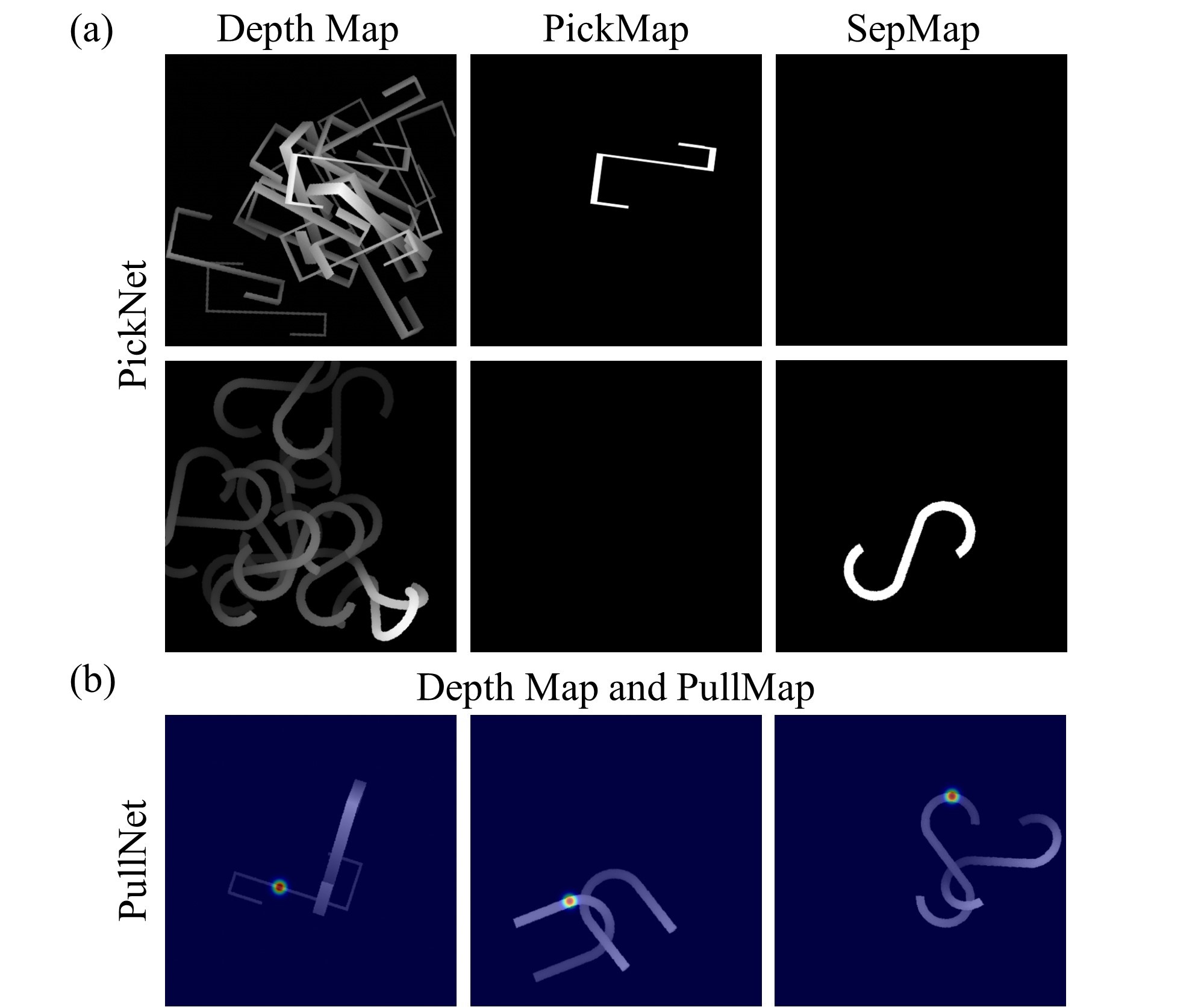}
    \caption{Ground truth labels for PickNet and PullNet.}
    \label{fig:data-gt}
\end{figure}

\begin{figure}[t]
    \centering
    \includegraphics[width=\linewidth]{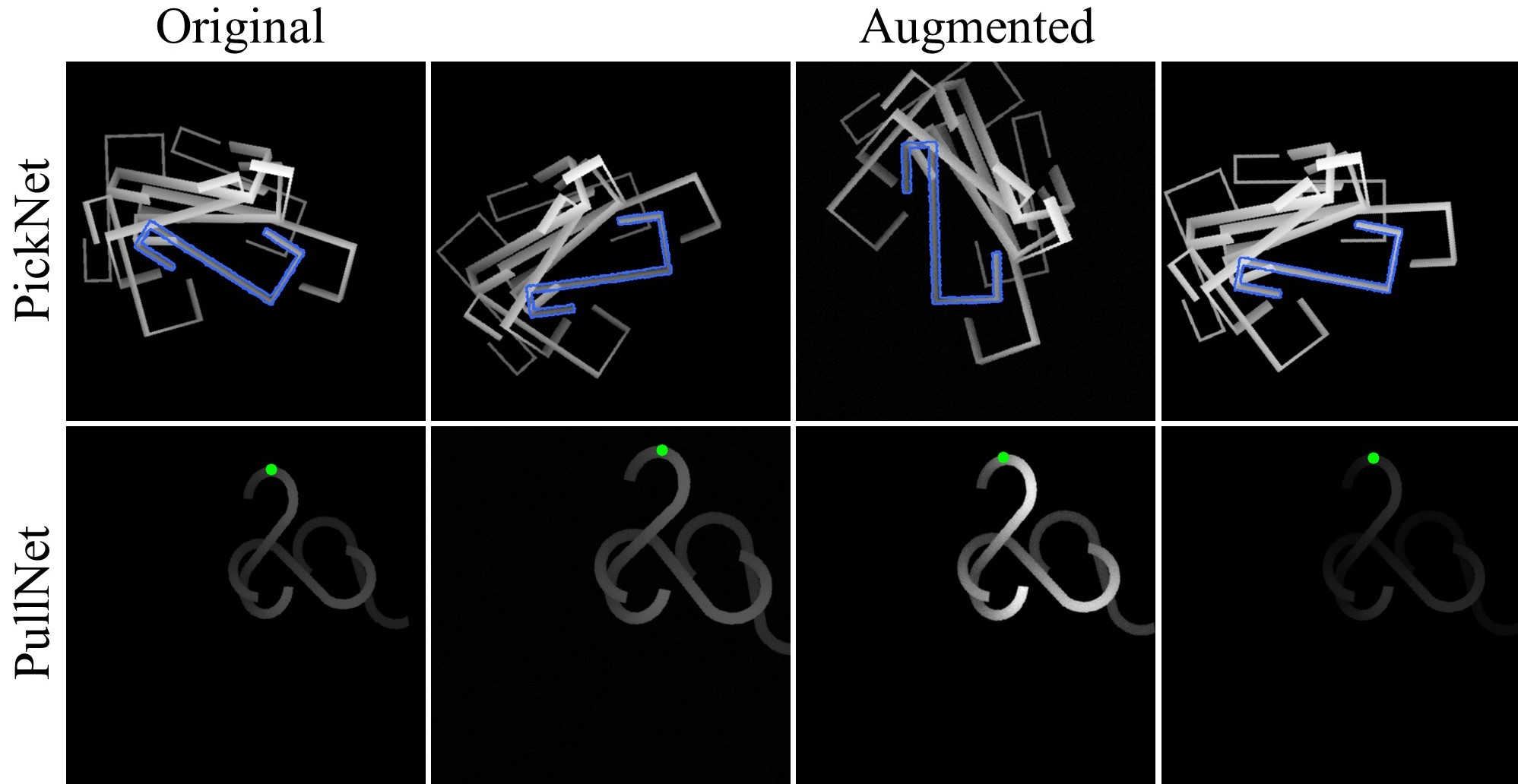}
    \caption{Augmented data for PickNet and PullNet. }
    \label{fig:data-aug}
\end{figure}

    \textit{1) Dataset:} The ground truth of a PullNet data sample is a single channel heatmap shown in Fig. \ref{fig:data-gt}(b). The position of pulling is encoded using Gaussian 2D while the direction of pulling is encoding by rotating the image so that the direction points to the right side in the image. We set the kernal of Gaussian 2D as 8. For data augmentation, we didn't apply rotations on PullNet dataset since we encodes the direction of pulling by rotating the image. We implement other image-based transformations as Table \ref{tab:data-aug} shows. Fig. \ref{fig:data-aug} also presented the augmented PullNet samples. Finally, we augmented the PullNet data 4X to 22,208.

    \textit{2) Training Details:} PullNet learns a mapping $f_\text{pull}: \mathbb{R}^{W \times H \times 3} \rightarrow \mathbb{R}^{W \times H}$. For the network architecture of PullNet, we use a ResNet-18 as the encoder, followed by a bi-linear upsampling layer pre-trained on ImageNet. We use the binary cross entropy Loss (BCE) as loss function. We train PullNet with a batch size of 2 using the Adam optimizer with a learning rate of 0.001 on a Nvidia GeForce RTX 3080 GPU. We finally select the weights from the 11-th epoch since it achieve the best performance.

\begin{table}[t]
\renewcommand\arraystretch{1}
\centering
\small
\caption{PickNet/PullNet Data Augmentations}
    \begin{tabular}{@{\extracolsep}lcc}
    \toprule
    \multirow{2.5}{*}{Augmentation Parameters} & \multicolumn{2}{c}{Amount}\\
    \cmidrule(lr){2-3}
    & PickNet & PullNet \\
    \midrule
    Additive Gaussian Noise & (0.0, 0.01*255) & (0.0, 0.01*255)\\
    Gamma Contrast & (0.5, 2.0) & (0.5, 2.0)\\
    Elastic Transformation & (1,1) & (1,1)\\
    Scale & (0.9,1.1) & (0.9,1.1)\\
    Shear & (-10,10) & (-10,10) \\
    Rotate & (-180,180) & - \\
    \bottomrule
    \end{tabular}
\label{tab:data-aug}
\end{table}  

\begin{figure*}[t]
    \centering
    \includegraphics[width=\linewidth]{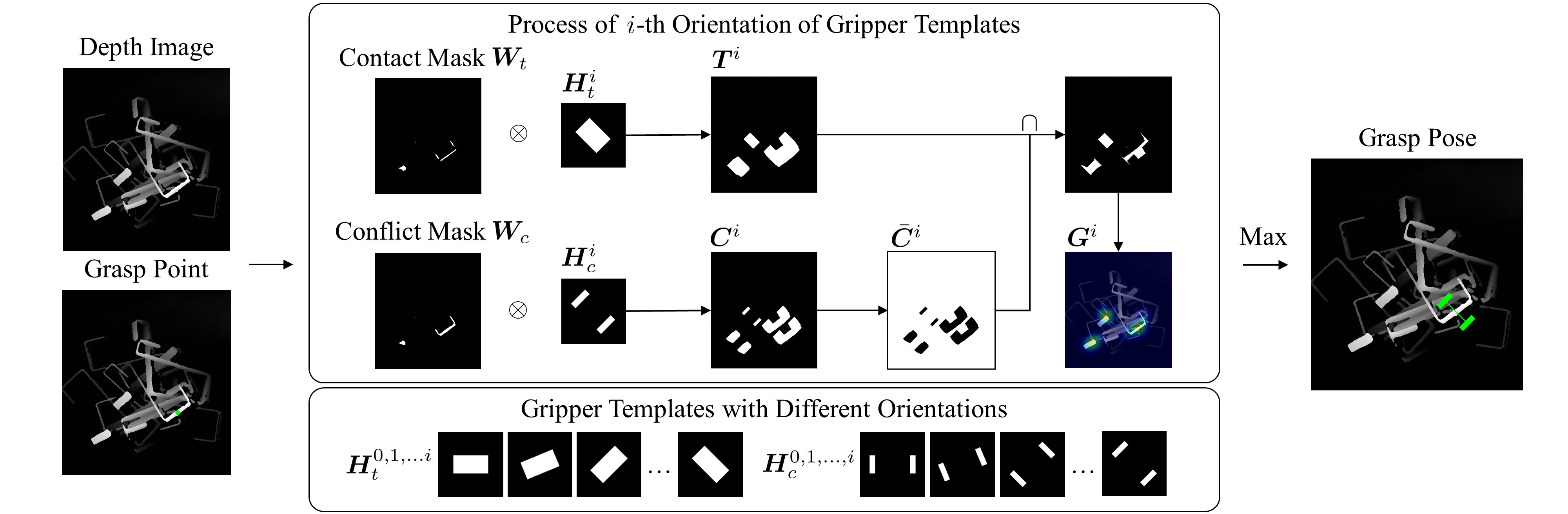}
    \caption{Grasp orientation detection.}
    \label{fig:grasp-detection}
\end{figure*}

\section{Experiments Details}
\subsection{Grasp Pose Detection}\label{subsec:exp-grasp}

    Fig. \ref{fig:grasp-detection} illustrates our method for detecting grasp orientation with a given grasp location. To detect collision-gree grasp orientation for already determined grasp location in the clutter, we revised the method Fast Graspability Evaluation (FGE). FGE constructs pixel-wise graspability scores with the input depth image by convoluting a template of contact areas and collision areas for the gripper. The output is a pixel location on the depth map and a rotation angle indicating the orientation of the parallel jaw gripper. Given an input of a depth image and the grasp pixel $p$, we construct the contact region $\bm{W}_t$ by the cross-section on the depth value of $p$, and the conflict region $\bm{W}_t$ by the cross-section on the depth value lower than that of $p$. Then, a set of gripper templates of opening $\bm{H}_t^{0,1,...}$ and closing $\bm{H}_t^{0,1,...}$ with different orientations are obtained. We convolute $\bm{W}_t$ with a template mask of a closing trajectory of gripper $\bm{H}^i_C$, and $\bm{W}_c$ with a template mask of two opening fingers $\bm{H}^i_C$ where $i$ denotes the different rotations for both gripper templates. Finally, we combine the results of convolution $\bm{T}^i$ and $\bar{\bm{C}}^i$ (bit-wise inversion of $\bm{C}^i$) and apply Gaussian Blur Filter. The output $\bm{G}_i$ with the highest pixel value denotes the index $i$ of the best grasp orientation. 

    For data collection in simulation, we develop an grasp detection algorithm \texttt{DetectGrasp()} (in Algorithm \ref{alg:supervisor-tangle}) where the goal is to detect the position and orientation of the grasp for an object with a known mask. This function is basically the same as the algorithm in Fig. \ref{fig:grasp-detection}. Instead of the grasp position as input, \texttt{DetectGrasp()} takes the depth image and the mask of the target as input, the contact mask $\bm{Wt}$ is revised as the cross section of the depth image masked with the target object. At the final ranking stage, the highest pixel location of $G_i$ with the orientation index $i$ are respectively the best grasp position and orientation. 


\begin{table}[t]
\renewcommand\arraystretch{1}
\centering
\small
\caption{Frequency of Unsuccessful Picking Attempts}
    \begin{tabular}{@{\extracolsep}clc}
    \toprule
    Method & Explanation & Frequency\\
    \midrule
    \multirow{2}{*}{PD} & (A) Grasps nothing & 4.7\% (22/462)\\
    & (B) Transport multiple objects & 8.8\% (41/462) \\
    \midrule
    \multirow{2}{*}{PDP} & (A) Grasps nothing & 5.0\% (21/422) \\
    & (B) Transport multiple objects & 5.9\% (25/422)\\
    \bottomrule
    \end{tabular}
\label{tab:data-augment}
\end{table}

\subsection{Failure Modes}

We divide the unsuccessful picking attempts as two types as follows: 
\begin{enumerate}[(A)]
    \item \textbf{The robot transports nothing to the goal bin.} The situation happens when the grasp poses are not correctly computed. PickNet produce a pixel location for our grasp detection algorithm to compute a 4-DoF grasp. Grasp failure occurs when each grasp orientation around the grasp location collided with the neighbor objects or the visual noise causes miscalculation in transforming 2D pixel locations to 3D locations, leading the gripper collides with the target, the neighbor objects or the bin walls. 
    \item \textbf{The robot transports multiple objects into the goal bin.} Sometimes due to the sensory noise, the correct locations of each object can not be presented from the depth map, e.g., parts of the objects are missing. Also, PickNet or PullNet sometimes make wrong predictions under some elusive entanglement situation or heavy occlusion. This may comes from the reality differ since the collision modelling of entanglement contact in the simulation still has difference with the real world. The physical execution of pulling sometimes cannot disentangle the objects due to insufficient pulling distance within the bin collisions. 
\end{enumerate}

    We present a total number of unsuccessful picking attempts through all seen and unseen objects for our policy PD and PDP as Table shows. The frequency is calculated by the number of unsuccessful picking attempts divided by the total number of attempts. Failure (A) occurs evenly in both policies. Our policy PDP with the entire workflow can significantly decrease the frequency of failure (B), showing the capabilities of disentangling objects. 

\subsection{Reasons of Using a Buffer Bin}

\begin{figure}[t]
    \centering
    \includegraphics[width=\linewidth]{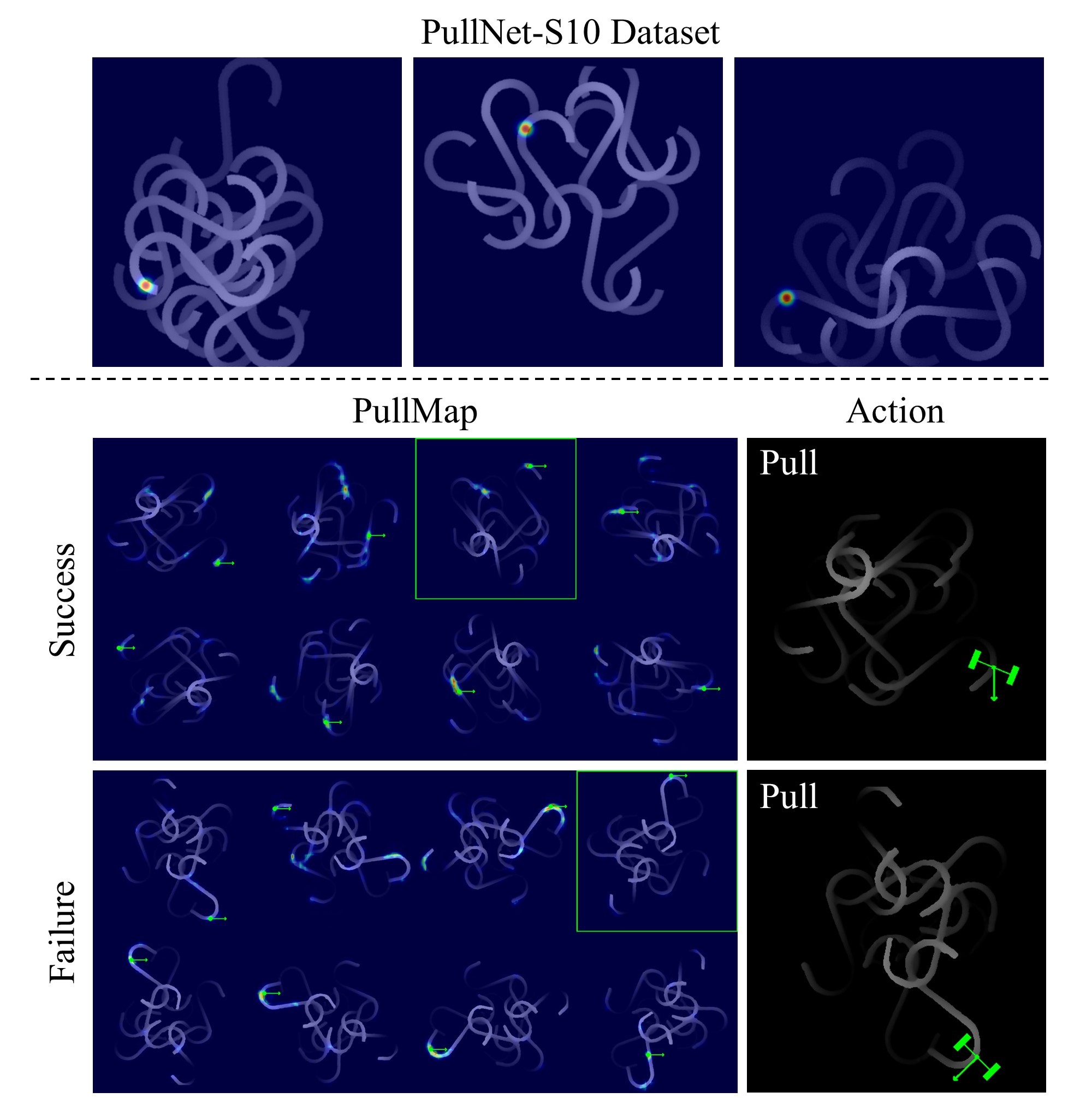}
    \caption{Dataset to train PullNet using 10 s-shaped objects and the prediction results.}
    \label{fig:pull-more}
\end{figure}

    We conducted a real-world experiment to test the pulling performance for different numbers of objects in the bin. We re-trained PullNet with the same architecture but with a different dataset. We collect the new dataset containing more than ten objects. The new dataset has 1000 augmented depth images and each image contains more than ten s-shaped objects as Fig. \ref{fig:pull-more} shows. We name the newly trained model PullNet-S10. To distinguish with the PullNet used in our policy, we evaluate the pulling success rates on the clutter with ten entangled s-shaped objects compared with those using our PullNet on two entangled s-shaped objects. The visualization results and numerical results are respectively shown in Fig. \ref{tab:pull-more} and Table \ref{tab:pull-more}. It demonstrates that the success rate of pulling under ten entangled objects is lower than that of two entangled objects. We guess the current self-supervised training manner might be unsuitable for predicting skillful manipulation strategies in complex and challenging environments. 
    
    Therefore, instead of directly pulling in the main bin, we leverage a buffer bin to first reduce the degrees of entanglement by dropping. The performance of PullNet under environments with less than five objects is significantly improved. Using a buffer bin can reduce the challenging entanglement phenomenon and dynamically disentangle the objects without visually planning precise skillful actions. Moreover, a buffer bin with fewer objects can avoid the challenging cases where motion planning of pulling actions sometimes collides with other objects. We observed some cases where the target was successfully pulled out but then entangled with the neighbor objects again. Other cases showed that the pulling distances were significantly constrained by the collision of a large number of objects. Therefore, we leverage a buffer bin to create an empty environment to plan motions for pulling and increase the success rates of pulling. 

\begin{table}[h]
\renewcommand\arraystretch{1}
\centering
\caption{Pulling Success Rates}
    \begin{tabular}{@{\extracolsep}lcc}
    \toprule
    & 2 Objects & 10 Objects\\
    \midrule
    Model & PullNet & PullNet-S10 \\
    Pulling Success Rate & 4/5 & 6/20\\
    \bottomrule
    \end{tabular}
\label{tab:pull-more}
\end{table}

\begin{figure*}[t]
    \centering
    \includegraphics[width=0.75\linewidth]{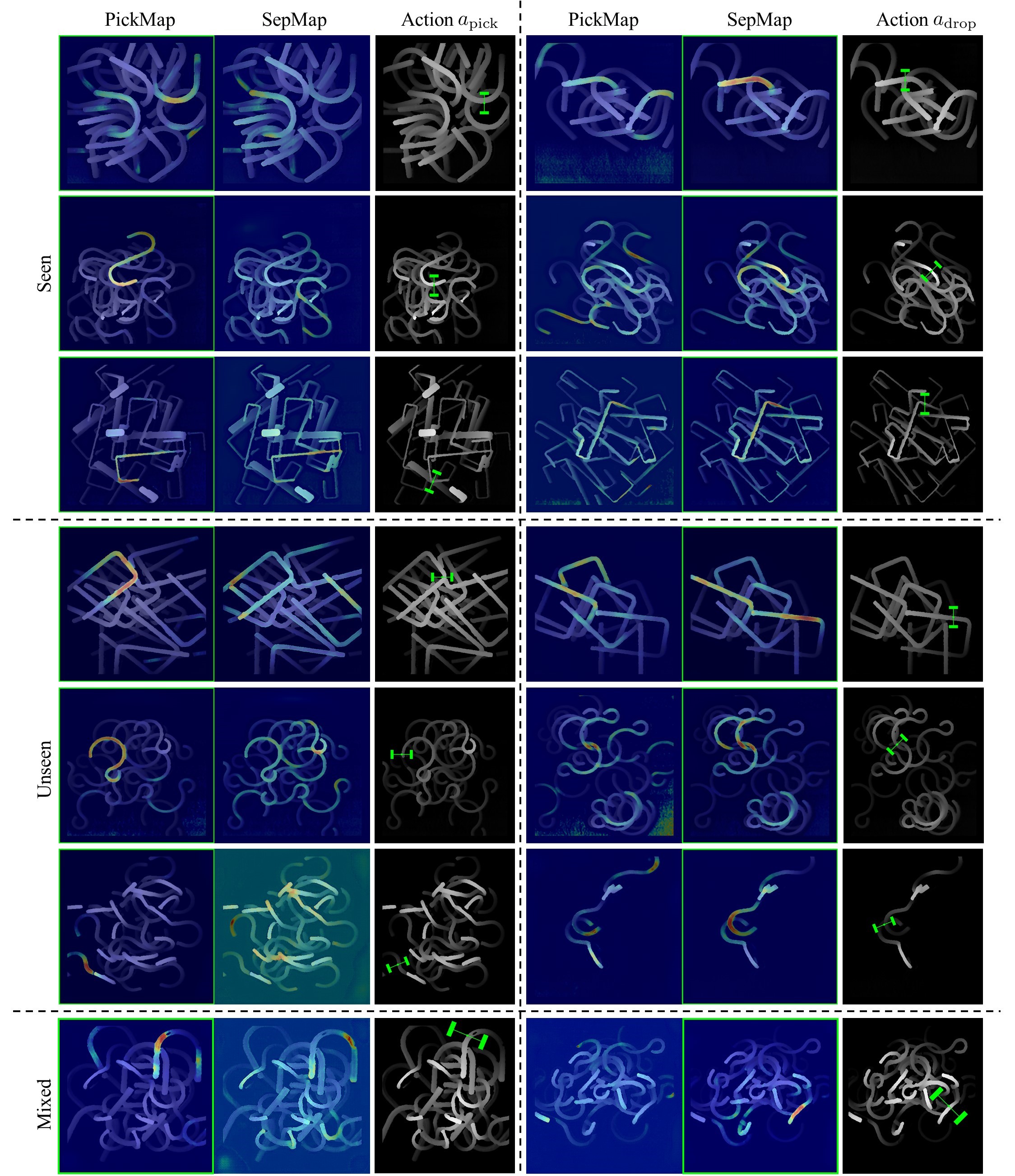}
    \caption{More visualized results using PickNet. }
    \label{fig:vis-pd}
\end{figure*}

\begin{figure*}[t]
    \centering
    \includegraphics[width=0.8\linewidth]{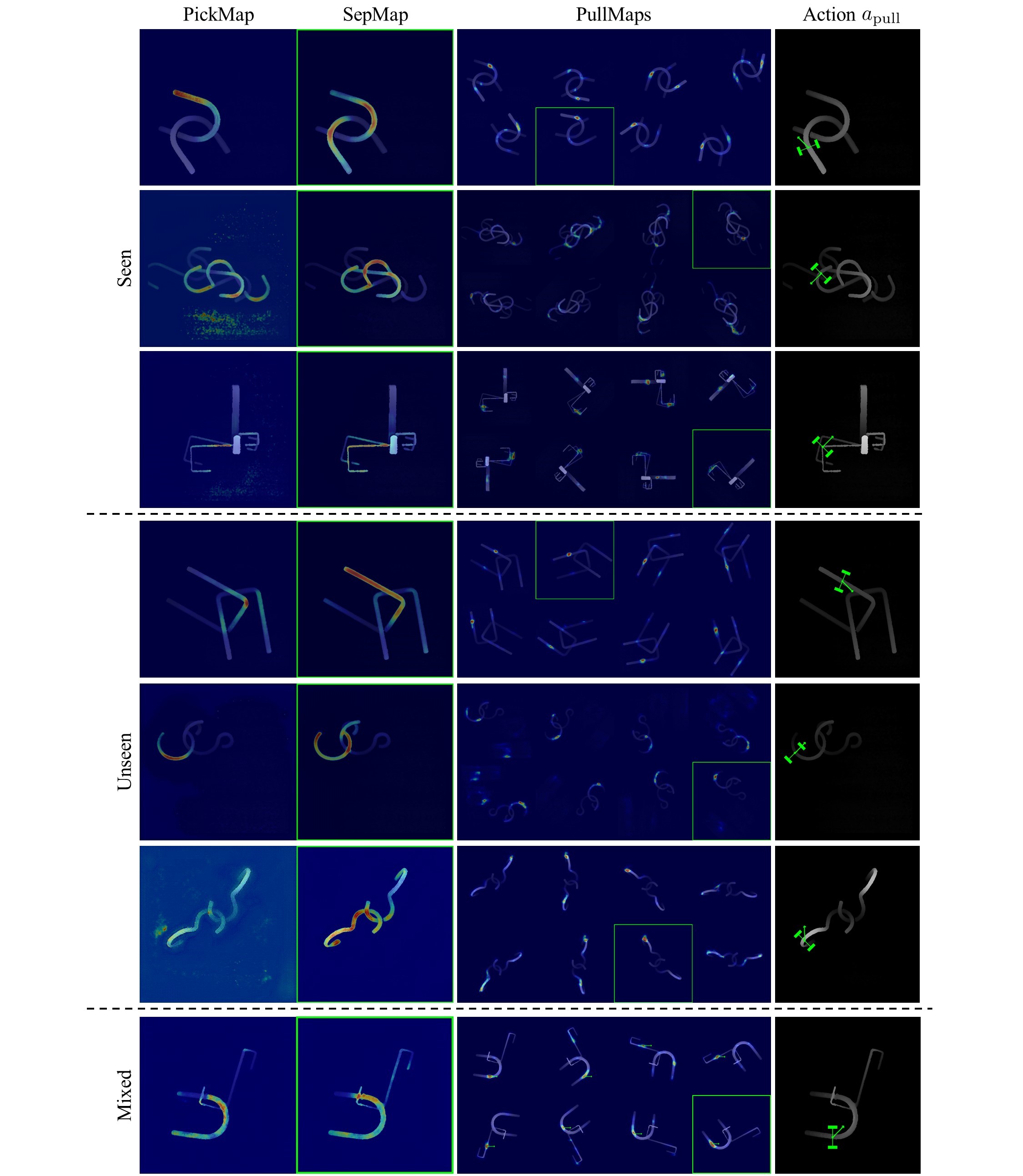}
    \caption{More visualized results where the bin contains only entangled objects using both PickNet and PullNet. }
    \label{fig:vis-pdp}
\end{figure*}
